\documentclass[manuscript,screen]{acmart}

\AtBeginDocument{%
  \providecommand\BibTeX{{%
    \normalfont B\kern-0.5em{\scshape i\kern-0.25em b}\kern-0.8em\TeX}}}

\setcopyright{acmcopyright}
\copyrightyear{2023}
\acmYear{2023}
\acmDOI{XXXXXXX.XXXXXXX}

\acmPrice{15.00}
\acmISBN{978-1-4503-XXXX-X/18/06}

\usepackage{bm}
\usepackage{bbm}
\usepackage{multirow}
\usepackage{makecell}

\begin{document}

\title{A Survey of Mix-based Data Augmentation: Taxonomy, Methods, Applications, and Explainability}

\author{Chengtai Cao}
\email{chengtcao2-c@my.cityu.edu.hk}
\orcid{0000-0003-3944-8358}
\affiliation{%
  \institution{City University of Hong Kong}
  \city{Hong Kong}
  \country{China}
}

\author{Fan Zhou}
\email{fan.zhou@uestc.edu.cn}
\authornote{Corresponding author: Fan Zhou}
\orcid{0000-0003-3944-8358}
\affiliation{%
  \institution{University of Electronic Science and Technology of China}
  \city{Chengdu}
  \country{China}
}

\author{Yurou Dai}
\email{yuroudai2@um.cityu.edu.hk}
\orcid{0000-0002-2553-6915}
\affiliation{%
  \institution{City University of Hong Kong}
  \city{Hong Kong}
  \country{China}
}

\author{Jianping Wang}
\email{jianwang@cityu.edu.hk}
\orcid{0000-0002-9318-1482}
\affiliation{%
  \institution{City University of Hong Kong}
  \city{Hong Kong}
  \country{China}
}

\author{Kunpeng Zhang}
\email{kpzhang@umd.edu}
\orcid{0000-0002-1474-3169}
\affiliation{%
  \institution{University of Maryland}
  \city{College Park}
  \country{USA}
}

\renewcommand{\shortauthors}{Chengtai Cao, Fan Zhou, Yurou Dai, Jianping Wang, and Kunpeng Zhang}

\begin{abstract}
Data augmentation (DA) is indispensable in modern machine learning and deep neural networks. The basic idea of DA is to construct new training data to improve the model's generalization by adding slightly disturbed versions of existing data or synthesizing new data. This survey comprehensively reviews a crucial subset of DA techniques, namely Mix-based Data Augmentation (MixDA), which generates novel samples by combining multiple examples. In contrast to traditional DA approaches that operate on single samples or entire datasets, MixDA stands out due to its effectiveness, simplicity, flexibility, computational efficiency, theoretical foundation, and broad applicability. We begin by introducing a novel taxonomy that categorizes MixDA into Mixup-based, Cutmix-based, and mixture approaches based on a hierarchical perspective of the data mixing operation. Subsequently, we provide an in-depth review of various MixDA techniques, focusing on their underlying motivations. Owing to its versatility, MixDA has penetrated a wide range of applications, which we also thoroughly investigate in this survey. Moreover, we delve into the underlying mechanisms of MixDA's effectiveness by examining its impact on model generalization and calibration while providing insights into the model's behavior by analyzing the inherent properties of MixDA. Finally, we recapitulate the critical findings and fundamental challenges of current MixDA studies while outlining the potential directions for future works. Different from previous related surveys that focus on DA approaches in specific domains (e.g., computer vision and natural language processing) or only review a limited subset of MixDA studies, we are the first to provide a systematical survey of MixDA, covering its taxonomy, methodology, application, and explainability. Furthermore, we provide promising directions for researchers interested in this exciting area. A curated list of reviewed methods can be found at https://github.com/ChengtaiCao/Awesome-Mix.
\end{abstract}

\ccsdesc[500]{Computing methodologies~Neural networks}
\ccsdesc[500]{Computing methodologies~Supervised learning by classification}
\ccsdesc[500]{Computing methodologies~Regularization}

\keywords{Data augmentation, mix strategies, generalization, machine learning.}

\received{20 February 2007}
\received[revised]{12 March 2009}
\received[accepted]{5 June 2009}

\maketitle

\section{Introduction}
\label{sec:introduction}
Deep learning (DL) has a transformative impact on diverse domains~\cite{lecun2015deep} due to its ability to learn expressive representation from data. As the complexity of the problems being addressed has increased, the architectures of deep neural networks (DNNs) have become increasingly sophisticated. However, DNNs are notorious for being data-hungry with millions, even billions of parameters (e.g., BERT~\cite{kenton2019bert}), making them prone to overfitting. 

Many innovations have been dedicated to making DNNs more data-efficient by developing improved network architectures. For example, convolutional neural networks (CNNs) undergo a remarkable evolution from AlexNet~\cite{krizhevsky2012imagenet} to Vision Transformers (ViTs)~\cite{dosovitskiy2020image}. Besides, various regularization techniques have been proposed to enhance the generalization capability of DNNs. Two notable examples are dropout~\cite{srivastava2014dropout} and batch normalization~\cite{ioffe2015batch}. Dropout randomly zeros out some activations during training to simulate an ensemble of sub-networks and prevent co-adaptation of neurons. On the other hand, batch normalization normalizes the activations by subtracting the batch mean and dividing by the batch standard deviation, which helps to stabilize the training process and improve convergence.

Data augmentation (DA), which involves increasing the size and the diversity of training data without explicitly collecting new examples, is commonly employed as a remedy to mitigate overfitting. DA methods aim to expand limited data and extract additional information, enhancing the overall model performance combined with advanced network architectures and existing regularization techniques. For instance, adding random noise into samples, as a simple DA method, can generate numerous new training samples to benefit model robustness. In the context of image data, label-invariant data transformations, such as random-cropping and horizontal-flipping~\cite{krizhevsky2012imagenet}, can boost model performance and robustness. Similarly, training a model with techniques like random erasing~\cite{zhong2020random} or Cutout~\cite{devries2017improved} can improve regularization. In the field of natural language processing (NLP), synonym replacement and random deletion~\cite{wei2019eda} are prevailing methods for augmenting textual data. Lastly, generative models, including variational auto-encoders (VAE)~\cite{kingma2014auto}, generative adversarial networks (GANs)~\cite{goodfellow2014generative}, and generative pre-trained transformers (GPT)~\cite{queiroz2020pre,yoo2021gpt3mix,anaby2020not,bayer2023data}, have gained popularity for DA due to their ability to generate an unlimited number of synthetic yet realistic samples.

This survey focuses on a burgeoning subfield of data augmentation -- Mix-based Data Augmentation (MixDA), which has aroused considerable research in recent years. In contrast to traditional DA methods operating on a single instance or entire dataset, MixDA creates virtual training data by combining multiple examples, thereby generating a great deal of training data without domain knowledge. For example, Mixup~\cite{zhang2018mixup} linearly interpolates input-output pairs from two randomly sampled training examples in a \textit{holistic} perspective. On the other hand, Cutmix~\cite{yun2019cutmix} cuts a patch from one image (source image) and then pastes it onto the corresponding region of another image (target image) from a \textit{locality} point of view. Subsequently, numerous improved versions of MixDA have been proposed, built upon the foundations of Mixup and Cutmix. These methods explore different aspects of data mixing, such as flexible mixing ratios, saliency guidance, and improved divergence, which form the basis for the taxonomy presented in this review. Owing to its versatility, MixDA has been successfully applied to a wide range of tasks, including semi-supervised learning, generative models, graph learning, and contrastive learning. Furthermore, several theoretical studies have been conducted to interpret and analyze MixDA from various perspectives.

\noindent \textbf{Importance of MixDA.} There are several key reasons why this technology has attracted widespread attention in the research community: (i) \textit{Effectiveness}: MixDA methods have consistently demonstrated superior performance compared to traditional DA techniques. By combining multiple samples, MixDA introduces a higher level of diversity in the training data, encouraging models to learn more robust and generalizable features; (ii) \textit{Simplicity} and \textit{Flexibility}: MixDA methods are relatively simple to implement and can be easily integrated into existing deep learning pipelines. They do not require extensive domain knowledge or complex transformations, making them accessible to a wide range of researchers and practitioners; (iii) \textit{Computational Efficiency}: Compared to other advanced data augmentation techniques, such as generative model based methods~\cite{kingma2014auto,goodfellow2014generative,yoo2021gpt3mix,anaby2020not,bayer2023data}, MixDA methods are computationally efficient. They do not require additional training or generation of synthetic samples, making them suitable for resource-constrained environments or large-scale datasets; (iv) \textit{Theoretical Foundations}: MixDA methods have solid theoretical foundations rooted in the principles of vicinal risk minimization, regularization, and calibration; and (v) \textit{Broad Applicability}: While MixDA methods have been primarily studied in the context of computer vision, their potential extends to other domains as well, such as speech recognition, natural language processing, and graph-structure data analysis. Given the rapid growth, effectiveness, versatility, and theoretical foundations of MixDA, it is evident that a comprehensive survey of this technique is both timely and necessary. This survey aims to provide a thorough overview of the foundations, methods, applications, and explainability of MixDA. By presenting our findings on the current state of MixDA, its challenges, and promising future research directions, we hope to illuminate the path for further advancements in this field.

\noindent \textbf{Related Surveys.} We clearly state the differences between our survey and related works to highlight its unique contributions. Several works have reviewed DA techniques~\cite{feng2021survey,yang2022image,shorten2019survey,wen2021time,zhao2021data,bayer2022survey,li2022data}, which are related to our work. However, these reviews primarily focus on applying DA methods in specific domains. For instance, \citet{feng2021survey} and \citet{li2022data} focus on DA methods in text data. Similarly, several surveys reviewing DA methods in other specific areas, such as image recognition~\cite{yang2022image,shorten2019survey}, time series learning~\cite{wen2021time}, graph learning~\cite{zhao2021data} and text classification~\cite{bayer2022survey}. Although there is some overlap between these works and our survey, such as \citet{bayer2022survey} providing a concise overview of DA methods for text classification, including some MixDA approaches for textual data, our study distinguishes itself by focusing exclusively on MixDA, which can be exploited in various domains (cf. Section~\ref{Sec_Application} for details). 
 
Among the related surveys, work~\cite{naveed2024survey} and work~\cite{lewy2022overview} are the most closely related to our survey. The former reviews the methods for \textit{image mix} and \textit{image deletion} while the latter reviews both mix augmentation and \textit{other} augmentation strategies. However, these works only provide a summary of a small subset of MixDA methods and have other focal points: \citet{naveed2024survey} review some deleting-based DA methods such as Random Erasing~\cite{zhong2020random} and Lewy et al.~\cite{lewy2022overview} discuss some cut-based DA techniques such as Cutout~\cite{devries2017improved} that erases a region of the input image with $0$ pixel value. Moreover, neither discusses the applications of MixDA across various domains, which is a crucial aspect of MixDA given its versatility. In contrast, our survey dedicates its full attention to MixDA, providing a comprehensive overview of its foundations, methods, and explainability. Most importantly, we also present a thorough review of MixDA applications, which previous studies have not covered. Furthermore, we present findings with respect to (w.r.t.) the current research and provide insight into remaining open challenges, as well as some promising future research directions. To our knowledge, this survey is the first comprehensive work to review MixDA techniques and summarize their wide spectrum of applications. In particular, we review more than $70$ MixDA methods and more than $8$ MixDA applications.

\noindent \textbf{Organization.} This survey is structured as follows. The overall picture of DA and MixDA is depicted in Section~\ref{Sec_Preliminary}, where we also provide a new classification of MixDA. Section~\ref{Sec_Methodology} systematically reviews existing methods in a more fine-grained taxonomy. Section~\ref{Sec_Application} investigates the important applications of MixDA, followed by the explainability analysis of MixDA in Section~\ref{Sec_Theory}. Moreover, the critical findings and challenges are presented in Section~\ref{Sec_Discussion}, which also outlines the potential research directions. Finally, we conclude this work in Section~\ref{Sec_Conclusion}.

\section{Preliminary}
\label{Sec_Preliminary}
\subsection{Data Augmentation (DA)}
The quantity and diversity of training data play a crucial role in the performance of machine learning models. To harness the full potential of machine learning approaches, numerous DA methods have been proposed to increase training data's size, variety, and quality by covering unexplored input space while ensuring the correctness of the associated labels. In addition to the differences in inherent augmentation techniques and their applicable domains~\cite{feng2021survey,yang2022image,shorten2019survey,wen2021time,zhao2021data,bayer2022survey,li2022data}, DA methods can be broadly categorized into three classes: (i) single-sample DA (SsDA), (ii) mix-based DA (MixDA), and (iii) generative-based DA (GenDA). Formally, SsDA approaches generate new instances $(\tilde{\mathbf{x}}_{i}, \tilde{\mathbf{y}}_{i})$ for the training sample $(\mathbf{x}_{i}, \mathbf{y}_{i})$ by applying transformations to the original sample while preserving its label:
\begin{equation}
    \tilde{\mathbf{x}}_{i} = \mathcal{A}_{\mathbf{\theta}}(\mathbf{x}_{i}), \ \tilde{\mathbf{y}}_{i} = \mathbf{y}_{i},
\end{equation}
where $\mathcal{A}_{\mathbf{\theta}}$ is a specific feature transformation operation (e.g., rotation) with parameters $\mathbf{\theta}$ (e.g., rotation angles). On the contrary, MixDA constructs synthetic data by combining multiple training examples, which can be formulated as:
\begin{equation}
\label{Eq_MixDA}
    \tilde{\mathbf{x}}_{i} = \mathcal{A}_{\mathbf{\theta}}(\mathbf{x}_{i}, \mathbf{x}_{j}), \ \tilde{\mathbf{y}}_{i} = \mathcal{A}_{\mathbf{\phi}}(\mathbf{y}_{i}, \mathbf{y}_{j}),
\end{equation}
where $\mathcal{A}_{\mathbf{\phi}}$ denotes the label combination scheme parameterized by $\mathbf{\phi}$. It is worth noting that Equation~\eqref{Eq_MixDA} illustrates the mixing of only two training examples for simplicity. However, extending the formulation to blend more samples is straightforward. Unlike SsDA and MixDA, which generate new data at the sample level, GenDA produces synthetic examples with labels based on the generative model $\mathcal{M}$. This model is trained or fine-tuned on the entire dataset $\mathcal{D}$:
\begin{align}
\label{Eq_GenDA}
     \mathcal{M} = \operatorname{Train}(\mathcal{M}, \mathcal{D}) & \ \text{or} \ \mathcal{M} = \operatorname{Fine-Tune}(\mathcal{M}, \mathcal{D}), \\
    (\tilde{\mathbf{x}}_{i}, \tilde{\mathbf{y}}_{i}) &= \mathcal{M}(),
\end{align}

GenDA can be easily distinguished from SsDA and MixDA, as the latter is at sample level. In contrast, GenDA requires training or fine-tuning a generative model on the entire dataset and then using the trained model to synthesize new instances. SsDA and MixDA have three main differences: (i) MixDA creates new training instances by combining multiple training samples, while SsDA only considers a single example; (ii) the feature transformation in SsDA is usually label-invariant, meaning that the newly generated instances have the same target label. In contrast, the label combination strategy $\mathcal{A}_{\mathbf{\phi}}$ in MixDA should be carefully designed to align with the feature mixing process; and (iii) most MixDA techniques are domain-agnostic and can be applied to various fields, whereas SsDA techniques are often domain-specific. Given the numerous advantages of MixDA, such as its effectiveness, flexibility, theoretical foundations, and broad applicability, we focus on this particular category of DA methods in this survey.

\subsection{Taxonomy of MixDA}
To better understand the landscape of MixDA methods, we propose a novel taxonomy that categorizes existing techniques into three main groups: (i) methods that mix training examples from a \textbf{global} perspective, represented by the pioneering work Mixup~\cite{zhang2018mixup}; (ii) approaches that construct new data through the lens of \textbf{locality}, represented by Cutmix~\cite{yun2019cutmix}; and (iii) other techniques that are based on the principle of mixing but cannot be simply grouped into the above two categories, such as mixing with data reconstruction and integrating multiple MixDA solutions. The rationale behind our proposed taxonomy is as follows. Mixup and its variants typically develop and apply a global mix scheme to all features. For instance, Mixup draws a mix ratio from a Beta distribution, and each feature in the created example is a linear combination of the corresponding features from two sampled training examples, encouraging the model to understand data \textit{globally}. In contrast, Cutmix and its adaptations intercept partial features from one instance and paste them on another, aiming to improve the model's \textit{localization} ability. Furthermore, several works exist that integrate multiple MixDA approaches or combine MixDA with other SsDA methods. For example, RandomMix~\cite{liu2022randommix} creates augmented data by sampling a mixing operation from a set of MixDA methods for each mini-batch. Similarly, AugMix~\cite{hendrycks2020augmix} constructs multiple versions for each sample using SsDA and then mixes them via MixDA techniques.

\section{MixDA Methods}
\label{Sec_Methodology}
\begin{table*}[ht]
    \caption{The commonly used MixDA benchmarks and learning tasks.}
    \label{Benchmark_Variant}
    \centering
    \begin{tabular}{r|c|c|l}
    \hline
    \textbf{Benchmark} & \textbf{Modality} & \textbf{Task} & \textbf{Article} \\
    \hline
    \multirow{4}{*}{CIFAR~\cite{krizhevsky2009learning}} & 
    \multirow{4}{*}{Image} 
        & Image Classification &\makecell[l]{~\cite{archambault2019mixup, chidambaram2022towards, guo2019mixup, dabouei2021supermix, baek2021gridmix, cascante2021evolving, greenewald2021k, chen2022stackmix, faramarzi2020patchup, berthelot2019understanding, bunk2021adversarially, feng2021shot, choi2022tokenmixup}\\
        ~\cite{harris2020fmix, hataya2022djmix, hendrycks2020augmix, hendrycks2022pixmix, hong2021stylemix, kim2021co, kim2020cut, kim2020puzzle, li2021feature, li2021boosting, liang2018understanding, liu2022augrmixat} \\
        ~\cite{mai2021metamixup, liu2022decoupled, liu2021unveiling, liu2018data, park2022unified, mangla2020varmixup, park2022saliency, pinto2022regmixup, qin2020resizemix, muhammad2021mixacm} \\
        ~\cite{venkataramanan2022teach, uddin2021saliencymix, venkataramanan2022alignmixup, verma2019manifold, rame2021mixmo, summers2019improved, sun2019patch, tokozume2018between, takahashi2018ricap} \\
        ~\cite{walawalkar2020attentive, yun2019cutmix, zhang2018mixup, zhang2021does, yu2021mixup, zhang2022m, zhu2020automix, yang2022recursivemix}} \\ \cline{3-4}
        & & Model Robustness Analysis & \makecell[l]{~\cite{chen2021guided, lamb2019interpolated, baena2022preventing, chou2020remix, kim2020puzzle, hendrycks2022pixmix, lee2020smoothmix, greenewald2021k, hong2021stylemix, faramarzi2020patchup, lee2020adversarial, lim2022noisy} \\ 
        ~\cite{verma2019manifold, wen2021combining, zhang2018mixup, liu2022decoupled, pang2020mixup, venkataramanan2022teach}} \\ \cline{3-4}
        & & Model Uncertainty Analysis &~\cite{wen2021combining, pinto2022regmixup} \\ \cline{3-4}
        & & Semi-Supervised Image Classification &~\cite{mai2021metamixup, rame2021mixmo, liu2022decoupled, verma2019interpolation, berthelot2019mixmatch, li2020dividemix, wei2020mixpul, li2022your, sun2022swapping} \\
    \hline
    \multirow{6}{*}{ImageNet~\cite{russakovsky2015imagenet}} &
    \multirow{6}{*}{Image} 
        & Image Classification &\makecell[l]{~\cite{harris2020fmix, guo2019mixup, hendrycks2020augmix, hendrycks2022pixmix, dabouei2021supermix, baek2021gridmix, cascante2021evolving, chen2022stackmix, hataya2022djmix, hong2021stylemix, choi2022tokenmixup, chen2022transmix, kim2021co} \\ 
        ~\cite{li2021boosting, li2020center, liang2018understanding, kim2020puzzle, li2021feature, liu2022tokenmix, liu2022augrmixat, liu2018data, lee2020smoothmix, liu2022decoupled, liu2021unveiling, mai2021metamixup} \\
        ~\cite{mangla2020varmixup, muhammad2021mixacm, park2022unified, park2022saliency, takahashi2018ricap, tokozume2018between, uddin2021saliencymix, qin2020resizemix, pinto2022regmixup, rame2021mixmo} \\
        ~\cite{zhang2018mixup, zhu2020automix, venkataramanan2022alignmixup, venkataramanan2022teach, verma2019manifold, yang2022recursivemix, yu2021mixup, yun2019cutmix}} 
         \\ \cline{3-4}
        & & Model Robustness Analysis &~\cite{zhang2018mixup, thulasidasan2019mixup, uddin2021saliencymix, chen2022transmix, kim2021co, venkataramanan2022alignmixup, cascante2021evolving, liu2022decoupled, pinto2022regmixup, lee2020adversarial, lim2022noisy} \\ \cline{3-4}
        & & Model Uncertainty Analysis &~\cite{thulasidasan2019mixup, chen2022transmix, venkataramanan2022alignmixup, wen2021combining, pinto2022regmixup} \\ \cline{3-4}
        & & Object Localization &~\cite{yun2019cutmix, kim2021co, venkataramanan2022alignmixup} \\ \cline{3-4}
        & & Pascal VOC Object Detection\ &\makecell[l]{~\cite{uddin2021saliencymix, chen2022transmix, cascante2021evolving, qin2020resizemix, hataya2022djmix, liu2021unveiling, venkataramanan2022teach, li2021feature, olsson2021classmix, li2020center, chu2020beyond, lee2021mix}} \\ \cline{3-4}
        & & MS-COCO Image Captioning\ & ~\cite{yun2019cutmix, chen2022transmix, yang2022recursivemix, yu2021mixup, cascante2021evolving, li2021boosting, venkataramanan2022teach, li2020center} \\ \cline{3-4}
    \hline
    \multirow{2}{*}{MNIST~\cite{lecun1998gradient}} &
    \multirow{2}{*}{Image} 
        & Image Classification &~\cite{guo2019mixup, chidambaram2022towards, mai2021metamixup, zhang2021does, harris2020fmix, baena2022preventing, zhu2020automix, greenewald2021k, beckham2019adversarial, berthelot2019understanding, feng2021shot} \\ \cline{3-4}
        & & Semi-Supervised Image Classification &~\cite{wei2020mixpul, li2022your} \\
    \hline
    CIFAR-C~\cite{hendrycks2019benchmarking} & Image & Model Robustness Analysis &~\cite{hendrycks2022pixmix, lee2020smoothmix, hendrycks2020augmix, rame2021mixmo, rame2021mixmo, liu2021unveiling, wen2021combining, pinto2022regmixup, laugros2020addressing, lim2022noisy, ren2022simple} \\
    \hline
    ImageNet-C~\cite{hendrycks2019benchmarking} & Image & Model Robustness Analysis &~\cite{hendrycks2022pixmix, lee2020smoothmix, hendrycks2020augmix, rame2021mixmo, rame2021mixmo, liu2021unveiling, wen2021combining, pinto2022regmixup, laugros2020addressing, lim2022noisy, ren2022simple} \\
    \hline
    \multirow{3}{*}{SVHN~\cite{netzer2011reading}} &
    \multirow{3}{*}{Image}
        & Image Classification &~\cite{verma2019manifold, baena2022preventing, mai2021metamixup, greenewald2021k, rame2021mixmo, faramarzi2020patchup, beckham2019adversarial, berthelot2019understanding, mangla2020varmixup, feng2021shot} \\ \cline{3-4} 
        & & Semi-Supervised Image Classification &~\cite{mai2021metamixup, verma2019interpolation, berthelot2019mixmatch, sun2022swapping} \\ \cline{3-4}
        & & Model Robustness Analysis &~\cite{lamb2019interpolated, lee2020adversarial, chen2021guided} \\ \cline{3-4}
    \hline
    \multirow{2}{*}{CUB200-2011~\cite{wah2011caltech}} &
    \multirow{2}{*}{Image}
        & Object Localization &~\cite{yun2019cutmix, baek2021gridmix, cascante2021evolving, li2020attribute, li2021boosting, liu2021unveiling} \\ \cline{3-4}
        & & Fine-Grained Image Classification &~\cite{huang2021snapmix, yu2021mixup, liu2021unveiling, liu2022decoupled} \\
    \hline 
    iNaturalist~\cite{van2018inaturalist} & Image
    & Unbalanced Image Classification &~\cite{chou2020remix, li2021boosting, liu2021unveiling} \\
    \hline
    FGVC-Aircraft~\cite{maji2013fine} & Image
    & Fine-Grained Image Classification &~\cite{huang2021snapmix, li2020attribute, li2021boosting, liu2021unveiling, liu2022decoupled} \\
    \hline
    UCI~\cite{asuncion2007uci} & Image
        & Tabular Data Classification &~\cite{zhang2018mixup, yu2021mixup, greenewald2021k} \\
    \hline
    MR~\cite{maas2011learning} & Image
        & Sentence Classification &~\cite{guo2020nonlinear, liu2021adversarial, guo2019augmenting} \\ 
    \hline
    TREC~\cite{pang2005seeing} & Image
        & Sentence Classification &~\cite{yoon2021ssmix, guo2020nonlinear, liu2021adversarial, kwon2022explainability, sawhney2022dmix, guo2019augmenting, jindal2020augmenting} \\
    \hline
    SST~\cite{socher2013recursive} & Text
        & Sentence Classification &~\cite{guo2020nonlinear, liu2021adversarial, kwon2022explainability, sawhney2022dmix, guo2019augmenting, jindal2020augmenting} \\
    \hline
    Subj~\cite{pang2004sentimental} & Text
    & Sentence Classification &~\cite{guo2020nonlinear, liu2021adversarial, kwon2022explainability, guo2019augmenting, jindal2020augmenting} \\ 
    \hline
    GLUE~\cite{wang2019glue} & Text
    & Natural Language Understanding &~\cite{yoon2021ssmix, yin2021batchmixup, sun2020mixup, zhang2022treemix} \\
    \hline
    Google Command~\cite{warden2018speech} & Audio
    & Audio Classification &~\cite{zhang2018mixup, harris2020fmix, kim2021co, li2021feature} \\
    \hline
    \end{tabular}
\end{table*}

In this section, we review a wide variety of MixDA strategies, which can be categorized into three groups based on our proposed taxonomy: (i) Mixup~\cite{zhang2018mixup} and its variants, (ii) Cutmix~\cite{yun2019cutmix} and its adaptations, and (iii) other MixDA methods. Specifically, in Section~\ref{Sub_Mixup}, we review Mixup-based methods. We begin by introducing the foundational work, Mixup, and then discuss its adaptations from various perspectives. Moving on to Section~\ref{Sub_Cutmix}, we start by presenting the influential work Cutmix. Subsequently, we examine its enhancements from different angles. Then, in Section~\ref{Sub_Others}, we review other mix-based methods that do not fit neatly into the two groups above. Finally, Section~\ref{Sub_Summary} discusses Mixup-based and Cutmix-based approaches, highlighting their strengths and weaknesses. To provide a comprehensive overview, we summarize the commonly used benchmarks and corresponding tasks in Table~\ref{Benchmark_Variant}, the characteristics of Mixup-based methods and Cutmix-based approaches in Table~\ref{Summary_Mixup} and Table~\ref{Summary_Cutmix}, respectively.

\subsection{Mixup-based Methods}
\label{Sub_Mixup}
\subsubsection{Mixup}
Mixup~\cite{zhang2018mixup} is the seminal work in MixDA, proposing a straightforward, data-independent, model-agnostic, effective, and efficient principle to construct new training instances. Mixup imposes an inductive bias on training distribution, assuming that linear interpolations of input pairs will result in convex combinations of their outputs, extending the training data distribution. From a regularization perspective, Mixup encourages the model to behave linearly between training examples. Specifically, the combination operation of Mixup is defined as follows:
\begin{equation}
\label{Eq_Mixup}
\tilde{\mathbf{x}} = \lambda \mathbf{x}_{i} + (1 - \lambda) \mathbf{x}_{j}, \ \tilde{\mathbf{y}} = \lambda \mathbf{y}_{i} + (1 - \lambda) \mathbf{y}_{j},
\end{equation}
where $(\mathbf{x}_{i}, \mathbf{y}_{i})$ and $(\mathbf{x}_{j}, \mathbf{y}_{j})$ are two data points randomly sampled from the original training distribution, and $(\tilde{\mathbf{x}}, \tilde{\mathbf{y}})$ is the generated instance. The targets $\mathbf{y}_{i}$ and $\mathbf{y}_{j}$ are usually represented as one-hot vectors, and $\lambda \in [0, 1]$ is a hyperparameter controlling the interpolation strength (also known as the mix ratio), which is typically sampled from a Beta distribution: $\lambda \sim \operatorname{Beta}(\alpha, \alpha)$, where $\alpha \in(0, \infty)$. Mixup shares similarities with SMOTE~\cite{chawla2002smote}, one of the most popular over-sampling approaches for imbalanced classification, as both methods synthesize new instances by interpolating between two existing instances. However, SMOTE performs interpolation between instances in minority classes and their nearest neighbors, while Mixup operates on randomly sampled instances. This difference arises because SMOTE primarily addresses the class imbalance problem, while Mixup is a general DA method.

\begin{figure}[!t]
\centering
\includegraphics[width=2.8in]{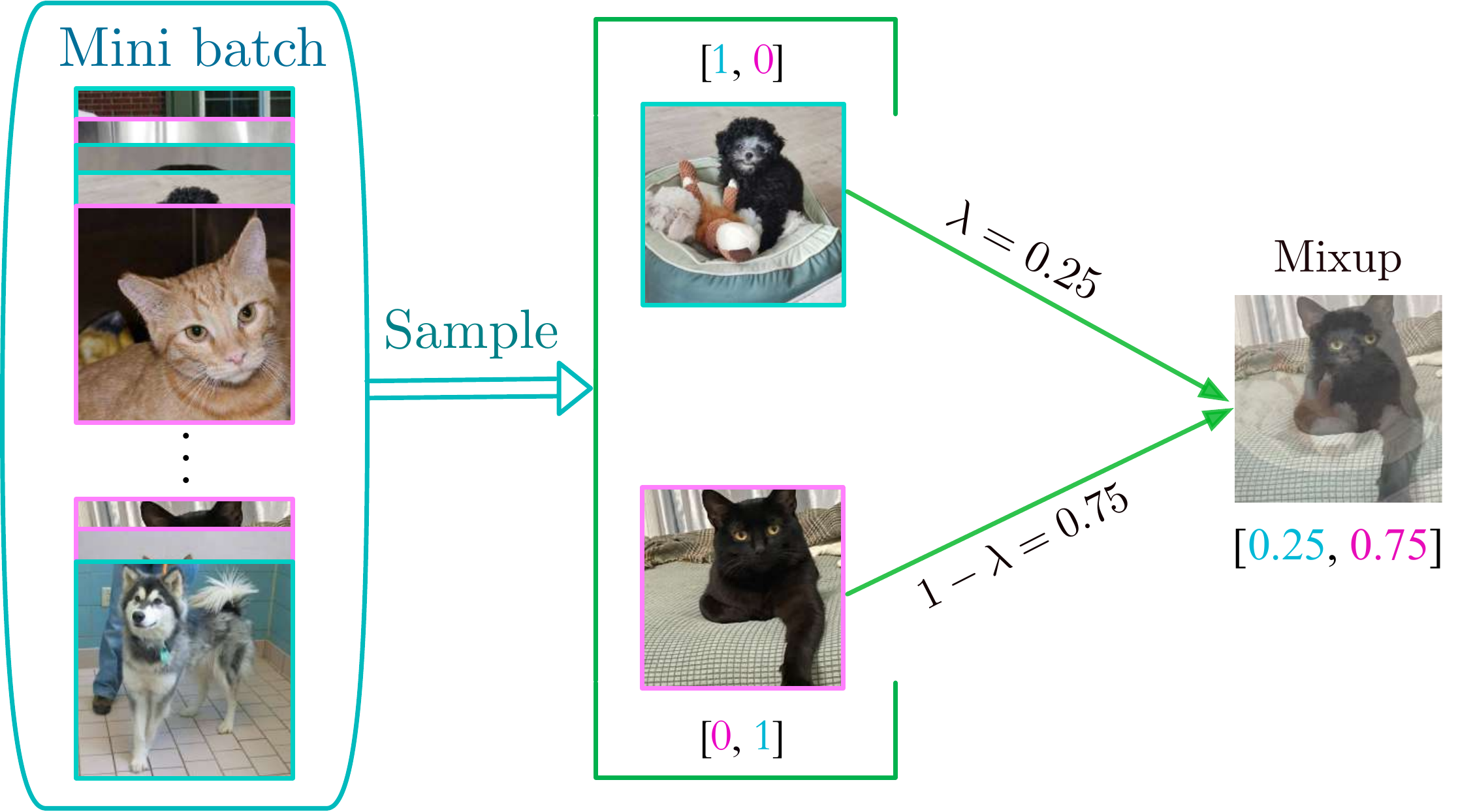}
\caption{The sketch of Mixup. Each pixel in the Mixup-generated image is a convex combination of the corresponding pixels from two randomly sampled images with the mix ratio $\lambda$.}
\label{Fig_Mixup}
\end{figure}

In practical implementation, Mixup operates on each mini-batch and synthesizes new training data using Equation~\eqref{Eq_Mixup}, introducing a small computation overhead. A sketch of Mixup is shown in Figure~\ref{Fig_Mixup}, where each pixel in generated images is a linear combination of corresponding pixels from sampled images, using a mixing ratio of $\lambda = 0.25$.

Concurrently with Mixup~\cite{zhang2018mixup}, \citet{tokozume2018learning} present a Between-Class (BC) learning strategy that composes synthetic sounds by mixing two origin samples and feeds the virtual data to the model. Unlike Mixup, which explicitly blends the target pairs, BC learning encourages the model to predict the mix ratio $\lambda$. Another critical difference between Mixup and BC learning is that the latter takes into account the properties of sound:
\begin{equation}
g=\frac{1}{1+10^{\frac{G_{i}-G_{j}}{20}} \cdot \frac{1-\lambda}{\lambda}}, \ \tilde{\mathbf{x}} = \frac{g\mathbf{x}_{i}+(1-g)\mathbf{x}_{j}}{\sqrt{g^{2}+(1-g)^{2}}},
\end{equation}
where $G_{i}$ and $G_{j}$ are the sound pressure levels [dB] of $\mathbf{x}_{i}$ and $\mathbf{x}_{j}$, respectively, and in essence, $\mathbf{x}_{i}:\mathbf{x}_{j}=\lambda:(1-\lambda)$. BC learning increases the diversity of training data and regularizes the feature space by maximizing the ratio of the between-class distance to the within-class variance (i.e., Fisher’s criterion). An extension of BC learning for image data is BC+~\cite{tokozume2018between}, which builds on the fact that CNNs process image data as waveforms.

\subsubsection{Mixing in Embedding Space}
Word2Vec, as introduced by \citet{mikolov2013efficient}, has revealed the intriguing property of linear relationships between word embedding. For instance, the equation "king - man + woman $\approx$ queen" demonstrates the ability to perform arithmetic calculations on word vectors, yielding semantically meaningful results. This linear compositionality of word embeddings has emerged as a desirable property for embedding spaces. Based on this observation, a natural approach is to linearly combine instances in the embedding space to generate augmented data. This approach leverages the inherent linearity in the embedding space to create new instances that preserve the semantic relationships encoded within the vectors.

Manifold Mixup~\cite{verma2019manifold} linearly combines the intermediate hidden representations of two inputs to utilize the feature space better, smoothing the decision boundaries at different levels of embedding space and reducing the intra-cluster distance. Like Mixup~\cite{zhang2018mixup}, the same linear composite of the corresponding output pair is constructed as the new supervision signal. More specifically, given a deep neural network with multiple layers, Manifold Mixup stochastically selects one layer and mixes the embeddings from two randomly sampled examples. The combined result then proceeds to the output layer. Due to the extra randomness in layer selection, compared with Mixup, the loss function in Manifold Mixup has an additional expectation term for optional layers. \citet{venkataramanan2022alignmixup} interpret Mixup from the deformation perspective and propose AlignMixup to further spatially align two sampled embeddings, enabling consistent interpolation in the feature space. To further smooth the decision boundaries and enhance model robustness, Noisy Feature Mixup (NFM)~\cite{lim2022noisy} injects noise when performing convex combinations of input-output pairs in the embedding space. This approach has been shown to achieve a favorable trade-off between performance on clean data and robustness against various types of data attacks.

Linear interpolation in the embedding space directly addresses the decision boundary issue and provides greater diversity than mixing solely in the data space. This approach generates augmented data with increased variation, leading to more effective regularization effects.

\subsubsection{Adaptive Mix Strategy}
\label{Sub_Ratio}
Mixup can be viewed as the imposition of "local linearity" constraints outside the data manifold, known as out-of-manifold regularization~\cite{guo2019mixup}. One potential limitation with Mixup is that the mix ratio, denoted as $\lambda$, obtained through a blind sampling process, may not be optimal. An original example in the data manifold assigned a soft label may conflict with its actual label. This phenomenon is known as "manifold intrusion."  

To adaptively generate a mix ratio, AdaMixUp~\cite{guo2019mixup} is proposed, which employs an auxiliary network to automatically determine a flexible combination scheme and designs a novel objective function to mitigate manifold intrusion. \citet{mai2021metamixup} apply the meta-learning paradigm to \textit{learn to mix}, which also aims to address the underfitting issue caused by manifold intrusion. Instead of using a predefined distribution for the mix policy, the introduced MetaMixUp~\cite{mai2021metamixup} dynamically determines the combination strategy in a data-adaptive manner. Specifically, MetaMixUp is a two-level framework in which a \textit{meta} model explores a new interpolation scheme that guides the \textit{main} model. AutoMix~\cite{liu2021unveiling} decomposes the mix training into two sub-tasks: mixed data generation with a Mix Block and mix classification. It unifies both tasks in an end-to-end framework to optimize the combination policy directly.

An unexpected phenomenon of Mixup training is that when combined with an ensemble model, calibration is undermined~\cite{wen2021combining} due to a trade-off between accuracy and calibration. Similar to manifold intrusion, this trade-off is because the soft target of mixed samples introduces an under-confidence issue, which is further aggravated by ensembles. To address this issue, motivated by the fact that some classes are more challenging to identify than others, CAMixup~\cite{wen2021combining}, instead of a policy from a Beta distribution, adjusts the mix ratio for each class based on its confidence and accuracy and only applies Mixup on challenging classes. Specifically, Mixup is not applied to the under-confident classes (where accuracy is higher than confidence) but is used to synthesize new data for the over-confident classes (where confidence is higher than accuracy) to reduce model confidence.

The linearity in Mixup constrains the diversity of augmented examples and leads to the manifold intrusion issue due to its simplicity, which can jeopardize the regularization effect. Nonlinear Mixup~\cite{guo2020nonlinear} is proposed to synthesize the input-output pairs adaptively. Unlike Mixup, which allocates the same mix ratio to each dimension of input, the input mixing strategy of Nonlinear Mixup uses a \textit{matrix} $\mathbf{M} \in \mathbb{R}^{W \times H}$ for the input $\mathbf{x_{i}} \in \mathbb{R}^{W \times H}$, where $W$ and $H$ are the width and height of images. Each value of $\mathbf{M}$ is independently sampled from the Beta distribution: $\tilde{\mathbf{x}} = \mathbf{M} \odot \mathbf{x}_{i} + (1 - \mathbf{M}) \odot \mathbf{x}_{j}$, where $\odot$ denotes element-wise multiplication. Note that Nonlinear Mixup differs from Cutmix (which will be detailed in the next sub-section) since the combination matrix of Cutmix~\cite{yun2019cutmix} is binary. When mixing target pair, Nonlinear Mixup projects one-hot label vector to $k$-dimensional representation $\mathbf{z} \in \mathbb{R}^{k}$ and the resulting matrix with $c$ rows (number of categories) and $k$ columns (dimension of label embedding) is then used to model targets. The combined instance $\tilde{\mathbf{x}}$ passes through a Policy Mapping module, enabling the label assignment of mixed $\tilde{\mathbf{x}}$ to be based on the synthetic input.

Adversarial Mixing Policy (AMP)~\cite{liu2021adversarial} constructs new examples with a perturbed mixing policy, which involves three steps: (i) using a random mixing ratio $\lambda$ to interpolate input-output pairs to create new data; (ii) imposing a slight adversarial perturbation to $\lambda$ to regenerate the mixed feature without changing the mixed target, introducing non-linearity to the model; and (iii) minimizing the loss function on the adjusted data to train the model.

Liu et al.~\cite{liu2022decoupled} identify an over-smoothing problem in Mixup that originates from the aimless mixing ratio and present a decoupled Mixup loss to achieve a good trade-off between discrimination and smoothness. Similarly, to adapt Mixup to class-imbalanced scenarios where the model is excepted to move the decision boundary towards the majority class, Remix~\cite{chou2020remix} is proposed. It disentangles the input and output combination coefficient to balance the generalization of minority and majority classes. The combination manipulation for input in Remix is the same as in Mixup. However, for the target combination, Remix sets a higher weight on the minority class to favor the label to the minority class.

In summary, employing a flexible combination strategy can effectively mitigate the manifold intrusion problem and address the associated issues of under-confidence and over-smoothing, thereby maximizing the effectiveness of MixDA.

\subsubsection{Sample Selection}
\label{Sub_SampleSelection}
When dissimilar samples are mixed in MixDA, the synthesis of out-of-distribution (OOD) samples can harm the model's performance. OOD samples during training can lead to decreased generalization and increased confusion for the model. To address this issue, careful sample selection is crucial.

In the context of semi-supervised named entity recognition (NER), Local Additivity based Data Augmentation (LADA)~\cite{chen2020local} generates novel examples by combining closed samples. LADA has two variants: Intra-LADA and Inter-LADA. Intra-LADA obtains new sentences by exchanging words within a single sentence and interpolating among these new sentences. Inter-LADA, on the other hand, combines different sentences to construct new data. Additionally, LADA proposes using a weighted mixture of random samples and $k$-nearest neighbor (KNN) samples to balance regularization and noise. For supervised neural machine translation, Continuous Semantic Augmentation (CSANMT)~\cite{wei2022learning} generates augmented data for each sample within a semantic adjacency region to cover sufficiently diverse synonymous representations. More specifically, an encoder is trained to assign a semantic neighborhood in the hidden space to each training sample, where tangent points within the neighborhood are semantically equivalent via tangential contrast. CSANMT proposes a Mixed Gaussian Recurrent Chain (MGRC) procedure to select a group of representations from the semantic neighborhood, and each of these is incorporated into the continuous hidden space. In the context of self-supervised learning~\cite{liu2021self}, \citet{zhang2022m} propose M-Mix to adaptively construct a succession of hard negatives, which can dynamically select multiple examples to mix and assign different weights based on similarity measurement or a learnable function. The regression problem differs from the classification task since interpolation in the continual target space may lead to arbitrarily incorrect labels. To address this issue, Hwang et al.~\cite{hwang2021mixrl} present RegMixup that learns a policy to determine which examples to mix based on the distance metrics using reinforcement learning.

Several other approaches share this motivation, such as decaying loss based on the distance between inputs (e.g., Local Mixup~\cite{baena2022preventing}), selecting similar inputs for combination (e.g., Pani~\cite{sun2019patch}), interpolating in hyperbolic space (e.g., HypMix~\cite{sawhney2021hypmix} and DMix~\cite{sawhney2022dmix}), employing local-emphasized and global-constrained sub-tasks (e.g., SAMix~\cite{li2021boosting}), and utilizing generative models (e.g., GenLabel~\cite{sohn2022genlabel}).

In conclusion, the adverse effects of OOD data in MixDA can be minimized by employing appropriate sample selection schemes during the mixing process. By carefully choosing samples that are more similar or share common characteristics, the synthesis of virtual examples through mixing becomes more plausible.

\subsubsection{Saliency \& Style Guidance}
\label{Sub_Saliency1}
As discussed by~\citet{huang1999statistics}, saliency information, which refers to the prominence or importance of specific features or objects within data, is informative and exhibits regularity in various contexts. This regularity in saliency information suggests that it can be leveraged to guide the Mixup process in data augmentation, potentially improving the effectiveness of the augmented instances.

SuperMix~\cite{dabouei2021supermix}, a supervised interpolation policy, is introduced to exploit the semantics of inputs for generating new training instances. SuperMix incorporates two additional loss functions: one for smoothing the mask matrices and the other for encouraging the sparsity of masks. StyleMix~\cite{hong2021stylemix} further distinguishes content features from style characteristics when mixing two input images, aiming to improve the variety of synthetic examples. Specifically, let $\mathcal{G}$ and $\mathcal{H}$ denote the pre-trained style encoder and style decoder, respectively. StyleMix obtains four features:
\begin{equation}
\label{Eq_StyleMix1}
    \mathbf{f}_{ii} = \mathcal{G}(\mathbf{x}_{i}), \ \mathbf{f}_{jj} = \mathcal{G}(\mathbf{x}_{j}), \
    \mathbf{f}_{ij} = \operatorname{AdaIN}(\mathbf{f}_{ii}, \mathbf{f}_{jj}), \ \mathbf{f}_{ji} = \operatorname{AdaIN}(\mathbf{f}_{jj}, \mathbf{f}_{ii}),
\end{equation}
where $\operatorname{AdaIN}$~\cite{huang2017arbitrary} adaptively normalizes features with the mean $\bm{\mu}(\cdot)$ and the standard deviation $\bm{\sigma}(\cdot)$:
\begin{equation}
\label{Eq_AdaIN}
    \operatorname{AdaIN}(\mathbf{f}_{ii}, \mathbf{f}_{jj}) = \bm{\sigma}(\mathbf{f}_{jj})(\frac{\mathbf{f}_{i i}-\bm{\mu}(\mathbf{f}_{i i})}{\bm{\sigma}(\mathbf{f}_{ii})})+\bm{\mu}(\mathbf{f}_{jj}).
\end{equation}
The mixed image is then generated by interpolating the four features:
\begin{equation}
\label{Eq_StyleMix2}
    \tilde{\mathbf{x}} = \mathcal{H}(t \mathbf{f}_{ii}+(1-\lambda_{t}-\lambda_{s}+t)\mathbf{f}_{jj} + (\lambda_{t}-t) \mathbf{f}_{ij}+(\lambda_{s}-t) \mathbf{f}_{ji}),
\end{equation}
where $\lambda_{t}$ and $\lambda_{s}$ are the content and style mix ratios drawn from the Beta distribution. Hyperparameter $t$ is constrained by $\max (0, \lambda_{t}+\lambda_{s}-1) \leq t \leq \min (\lambda_{t}, \lambda_{s})$ and the combined target of the mixed image is:
\begin{equation}
\label{Eq_StyleMix3}
    \mathbf{y}_{c} = \lambda_{t} \mathbf{y}_{i} + (1 - \lambda_{t})\mathbf{y}_{j}, \ \mathbf{y}_{s} = \lambda_{s} \mathbf{y}_{i} + (1 - \lambda_{s})\mathbf{y}_{j}, \
    \tilde{\mathbf{y}} = \lambda \mathbf{y}_{c} + (1 - \lambda) \mathbf{y}_{s}.
\end{equation}

Inspired by the observation that the visual domain is associated with image style (e.g., abstract painting vs. landscape), Mixstyle~\cite{zhou2021domain} probabilistically combines instance-level feature statistics of training data. Due to the modular nature of CNNs, style information is modeled in the bottom layers, where the proposed style-mixing operation occurs. Combining styles of the training data may implicitly synthesize new domains, enhancing the domain diversity and further the generalization of the model. Specifically, MixStyle follows the paradigm of mini-batch training, i.e., shuffling and mingling procedures are conducted along the batch dimension:
\begin{equation}
\gamma_{\text{mix}}=\lambda \bm{\sigma}(\mathbf{x})+(1-\lambda) \bm{\sigma}(\hat{\mathbf{x}}), \ \beta_{\text{mix}} =\lambda \bm{\mu}(\mathbf{x})+(1-\lambda) \bm{\mu}(\hat{\mathbf{x}}), \
\operatorname{MixStyle}(\tilde{\mathbf{x}})=\gamma_\text{mix} \frac{\mathbf{x}-\bm{\mu}(\mathbf{x})}{\bm{\sigma}(\mathbf{x})}+\beta_\text{mix}, 
\end{equation}
where $\hat{\mathbf{x}}$ is the batch reference representation obtained by shuffling the sampled mini-batch from two domains. The mean vector $\bm{\mu}(\cdot)$ and standard deviation vectors $\bm{\sigma}(\cdot)$ are calculated across the spatial dimension within each channel of each instance, similar to batch normalization~\cite{ioffe2015batch}. 

\citet{huang2017arbitrary} demonstrate that positional moments (i.e., mean $\bm{\mu}$ and standard deviation $\bm{\sigma}$) and instance moments can approximately represent shape and style information, which are usually discarded to accelerate and stabilize training. In contrast, Moment Exchange (MoEx)~\cite{li2021feature} uses this instructive information to generate new samples. To be specific, MoEx switches the moments of an example with those of the other:
\begin{equation}
\label{Eq_MoEx1}
    \mathbf{h}_i^{(j)}=\bm{\sigma}_j \frac{\mathbf{h}_i-\bm{\mu}_i}{\bm{\sigma}_i}+\bm{\mu}_j,
\end{equation}
where $\mathbf{h}_i^{(j)}$ is the resulting feature for the $i$-th input's representation $\mathbf{h}_{i}$ mixed with the $j$-th input's representation $\mathbf{h}_{j}$. This combined feature then passes through the network to calculate loss w.r.t. labels $\mathbf{y}_{i}$ and $\mathbf{y}_{j}$.

For natural language understanding tasks, \citet{park2022calibration} propose a combination scheme to calibrate a pre-trained language model based on saliency information and the Area Under the Margin (AUM) statistic. Training data is first divided into two groups -- a set of samples that are easy to distinguish and a batch of ambiguous samples that are difficult to learn by computing the AUM of each instance. Then, a Mixup process is executed to blend these two sets based on the saliency signal proxied by the gradient norm, reconciling the learning complexity and sample confidence for model calibration. Similarly, explainable artificial intelligence (XAI)~\cite{kwon2022explainability} explicitly extracts the weight of manipulated words and calculates the mixed label based on their importance. Furthermore, to circumvent the intense computation in saliency detectors, TokenMixup~\cite{choi2022tokenmixup} leverages attention map to guide token-level data augmentation by optimally matching sample pairs, achieving a $15\times$ speed-up. SciMix~\cite{sun2022swapping} is proposed to generate new samples with many characteristics not from their semantic parents by teaching a StyleGAN generator to replace global semantic content from other samples into image backgrounds.

Saliency and style information can be leveraged to deduce a content-aware combination policy, providing guidance for the mix process and generating more reasonable augmented data.

\subsubsection{Diversity in Mixup}
\label{Sub_Diversity1}
Mixup has specific constraints that can limit the diversity of the augmented data. The primary constraints are linear interpolation within the same batch and blending only two examples.

BatchMixup~\cite{yin2021batchmixup} breaks the same-batch rule by interpolating all mini-batches, enlarging the space of Mixup-ed examples. $K$-Mixup~\cite{greenewald2021k} produces more augmented data through perturbing $K$ instances in the direction of other $K$ samples and interpolates under the Wasserstein metric. Similarly, \citet{jeong2021observations} present a novel $K$-image mixing paradigm based on the stick-breaking process under Dirichlet prior. To interpolate an arbitrary number of samples, MultiMix~\cite{venkataramanan2022teach} assigns one vector from a Dirichlet distribution to each sample and implements the mix operation in the network's last layer. Instead of single output in Mixup, MixMo~\cite{rame2021mixmo} is presented with multi-input multi-output sub-modules for combining multiple samples. Specifically, inputs are passed through shared encoding layers and then mingled. The mixed representations are fed to multiple output blocks corresponding to multiple labels. To further enlarge the space of augmented data, \citet{hendrycks2022pixmix} present PixMix to combine origin training data with structurally complex images, which include fractals from DeviantArt\footnote{https://www.deviantart.com} and feature visualizations OpenAI Microscope\footnote{https://openai.com/blog/microscope}. 

To summarize, expanding the range of samples to mix and combining more examples during the convex interpolation process are effective strategies to enhance the variety of the constructed data in Mixup-based data augmentation.

\subsubsection{Miscellaneous Mixup Methods}
In addition to the previously described methods, there are other approaches to improve Mixup from other perspectives.

To obtain better \textit{robustness}, a guided interpolation framework (GIF)~\cite{chen2021guided} is developed to leverage the metainformation from the previous epochs for interpolation guidance. Compared with the conventional Mixup, GIF generates more attackable data (i.e., $\lambda=0.5$). Besides, GIF tranquilizes the linear behavior between classes, which benefits standard training but is not favorable to adversarial training for robustness, encouraging the model to predict invariably in a cluster. Through the lens of the \textit{exploration-exploitation} dilemma in reinforcement learning, Mixup constantly explores the data space. To achieve a pleasurable trade-off between exploration and exploitation, Mixup Without hesitation (MWh)~\cite{yu2021mixup} is proposed to divide a set of mini-batch examples into $3$ parts: (i) the standard Mixup, (ii) the combination of Mixup and basic data augmentation, and (iii) a damped Mixup. Mixup has been proved to be equivalent to discovering Euclidean \textit{barycenters} to boost the performance of the classifier~\cite{zhu2020automix}. Additionally, motivated by the connection between cross-entropy loss and Kullback–Leibler (KL) divergence, AutoMix~\cite{zhu2020automix} constructs barycenter samples with a generative model where OptTransMix utilizes Wasserstein distance instead of the Euclidean metric to define the mixing policy with the optimal transport-based transformation technique. RegMixup~\cite{pinto2022regmixup} trains the model on data distribution, which combines the original data and the Mixup vicinal approximation to the data distribution instead of using any of them as the exclusive training objective as in standard empirical risk minimization or original Mixup. This \textit{additional Mixup regularization term} to the standard loss function provides improved accuracy and uncertainty estimation when evaluating models under OOD data and numerous covariate shifts. Though original Mixup tends to obtain high-entropy models, it cannot separate in-distribution instances from out-distribution ones.

\begin{table*}[ht]
    \caption{Summary of Mixup-based DA methods}
    \label{Summary_Mixup}
    \centering
    \begin{tabular}{r|c|l}
    \hline
    \textbf{Adaptation} & \textbf{Method} & \textbf{Characteristic} \\
    \hline
    \multirow{3}{*}{Mixing in Embedding Space}
    & Manifold Mixup~\cite{verma2019manifold} & Mix hidden representations \\
    & AlignMixup~\cite{venkataramanan2022alignmixup} & Spatially align embeddings when mixing \\
    & NFM~\cite{lim2022noisy} & Inject noise in embedding space \\
    \hline
    \multirow{8}{*}{Adaptive Mix Strategy}
    & AdaMixUp~\cite{guo2019mixup} & Auxiliary network determines combination \\
    & MetaMixUp~\cite{mai2021metamixup} & Meta model explores interpolation scheme \\
    & AutoMix~\cite{liu2021unveiling} & Decompose mix training into sub-tasks \\
    & CAMixup~\cite{wen2021combining} & Adjust mix ratio based on class confidence and accuracy \\
    & Nonlinear Mixup~\cite{guo2020nonlinear} & Input mix strategy is a matrix \\
    & AMP~\cite{liu2021adversarial} & Adversarial perturbation to feature mixing \\
    & Decoupled Mixup~\cite{liu2022decoupled} & Decouple Mixup loss to discrimination and smoothness \\
    & Remix~\cite{chou2020remix}  & Disentangle input and output combination coefficient \\
    \hline
    \multirow{10}{*}{Sample Selction}
    & LADA~\cite{chen2020local} & Mix with closed samples \\
    & CSANMT~\cite{wei2022learning} & Mix with semantic neighborhood \\
    & M-Mix~\cite{zhang2022m} & Dynamically mix based on similarity \\
    & RegMixup~\cite{hwang2021mixrl} &  Learn policy to select examples to mix \\
    & Local Mixup~\cite{baena2022preventing} & Loss decayed with input distance \\
    & Pani~\cite{sun2019patch} & Mix with similar samples \\
    & HypMix~\cite{sawhney2021hypmix} & Interpolate in hyperbolic space \\
    & DMix~\cite{sawhney2022dmix} & Select samples based on embedding diversity \\
    & SAMix~\cite{li2021boosting} & Decompose Mixup into local and global sub-tasks \\
    & GenLabel~\cite{sohn2022genlabel} & Learn class-conditional data distribution \\
    \hline
    \multirow{6}{*}{Saliency \& Style Guidance}
    & SuperMix~\cite{dabouei2021supermix} & Exploit input semantics \\
    & StyleMix~\cite{hong2021stylemix} & Discern content and style when mixing \\
    & Mixstyle~\cite{zhou2021domain} & Combine instance-level feature statistics \\
    & MoEx~\cite{li2021feature} & Leverage positional moments to generate samples\\
    & TokenMixup~\cite{choi2022tokenmixup} & Attention map guides token-level augmentation \\
    & SciMix~\cite{sun2022swapping} & Replace semantic content into backgrounds \\
    \hline
    \multirow{5}{*}{Diversity}
    & BatchMixup~\cite{yin2021batchmixup} & Interpolate across mini-batches \\
    & $K$-Mixup~\cite{greenewald2021k} & Interpolate $K$ instances with other $K$ samples \\
    & MultiMix~\cite{venkataramanan2022teach} & Mix scheme from Dirichlet distribution \\
    & MixMo~\cite{rame2021mixmo} & Multi-input multi-output modules combine samples \\
    & PixMix~\cite{hendrycks2022pixmix} & Combine training data with complex images \\
    \hline
    \multirow{4}{*}{Others}
    & GIF~\cite{chen2021guided} & Leverage previous epochs for interpolation \\
    & MWh~\cite{yu2021mixup} & Balance exploration and exploitation \\
    & AutoMix~\cite{zhu2020automix} & Wasserstein distance defines mix policy \\
    & RegMixup~\cite{pinto2022regmixup} & Combine original data and Mixup approximation \\
    \hline
    \end{tabular}
\end{table*}

\subsection{Cutmix-based Methods}
\label{Sub_Cutmix}
\subsubsection{Cutmix}
\begin{figure}[!t]
\centering
\includegraphics[width=2.8in]{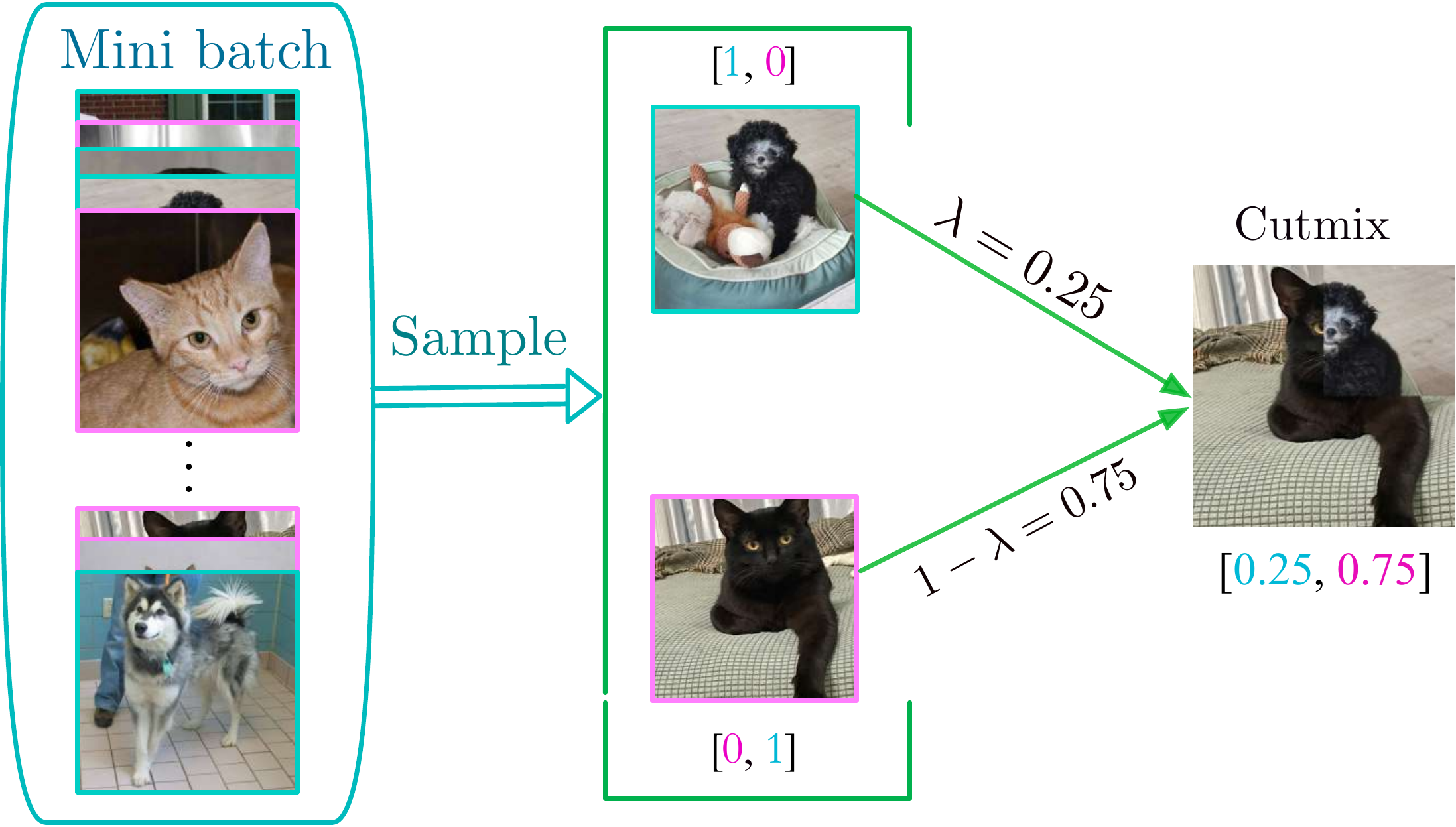}
\caption{An example of Cutmix with mix ratio $\lambda=0.25$, in which $1/4$ area of the dog image (upper right) is cut and pasted onto the corresponding location of the cat (upper right).}
\label{Fig_Cutmix}
\end{figure}

Inspired by Cutout~\cite{devries2017improved}, Random Erasing~\cite{zhong2020random}, and Cutoff~\cite{shen2020simple} that respectively randomly cut out a region (e.g., a rectangle) of the input images, replace the pixels of the selected area with random values, and remove part of the information within an input sentence, \citet{yun2019cutmix} design the Cutmix scheme which replaces a region of one image with the corresponding patch of the other sample, and blends their targets proportionally to the area of the combined examples. Compared with Cutout, Random Erasing, and Cutoff, each dimension of the Cutmix generated sample is informative, accelerating the training process. Besides, the features from the other sample upgrade the localization ability of the model by encouraging the model to identify the feature from a partial perspective. Specifically, Cutmix constructs new training samples by:
\begin{equation}
\label{Eq_Cutmix}
\tilde{\mathbf{x}} = \mathbf{M} \odot \mathbf{x}_{i} + (1 - \mathbf{M}) \odot \mathbf{x}_{j}, \ \tilde{\mathbf{y}} = \lambda \mathbf{y}_{i} + (1 - \lambda) \mathbf{y}_{j},
\end{equation}
where $\mathbf{M}$ is a \textit{binary} mask indicating the cut-and-paste area of the images. As with Mixup~\cite{zhang2018mixup}, the mix ratio $\lambda$ is drawn from a Beta distribution. To obtain the binary mask $\mathbf{M}$, Cutmix generates a bounding box by uniformly sampling a rectangle with height $\sqrt{\lambda}H$ and width $\sqrt{\lambda}W$. The binary matrix is determined by setting the element within the bounding box as 1 and 0 otherwise. A sketch of Cutmix is shown in Figure~\ref{Fig_Cutmix}, where the upper right $1/4$ (i.e., $\lambda=0.25$) patch of the dog image is cut and then pasted onto the corresponding region of the cat image. 

In parallel with Cutmix, work~\cite{summers2019improved} expands the space of mixed examples and investigates a dozen cut-and-paste policies. Taking $\textit{Vertical Concat}$ as an example, it vertically integrates the top $\lambda$ part of image $\mathbf{x}_{i}$ with the bottom $(1 - \lambda)$ region of image $\mathbf{x}_{j}$. Not limited to only combining two instances, random image cropping and patching (RICAP) blends $4$ images~\cite{takahashi2018ricap}. RICAP draws two parameters $a$ and $b$ from the uniform distribution: $a \sim \operatorname{Unif}(0, W), b \sim \operatorname{Unif}(0, H)$ and the cropping size $(w_{i}, h_{i})$ for $i$-th image is automatically obtained via $w_{1}=w_{3}=a$, $w_{2}=w_{4}=W-a$, $h_{1}=h_{2}=b$, and $h_{1}=h_{4}=H-b$. For $i$-th image, the coordinate of the upper left corner $[m_{i}, n_{i}]$ of the cropped areas is decided by $m_{i} \sim \operatorname{Unif}(0, W-w_{i})$ and $n_{i} \sim \operatorname{Unif}(0, H-h_{i})$. Similar to Cutmix, the mingled target is defined as mixing their one-hot vectors with the proportional to their areas in the generated image.

\subsubsection{Integration with Saliency Information}
\label{Sub_Saliency2}
\begin{figure}[!t]
\centering
\includegraphics[width=2.8in]{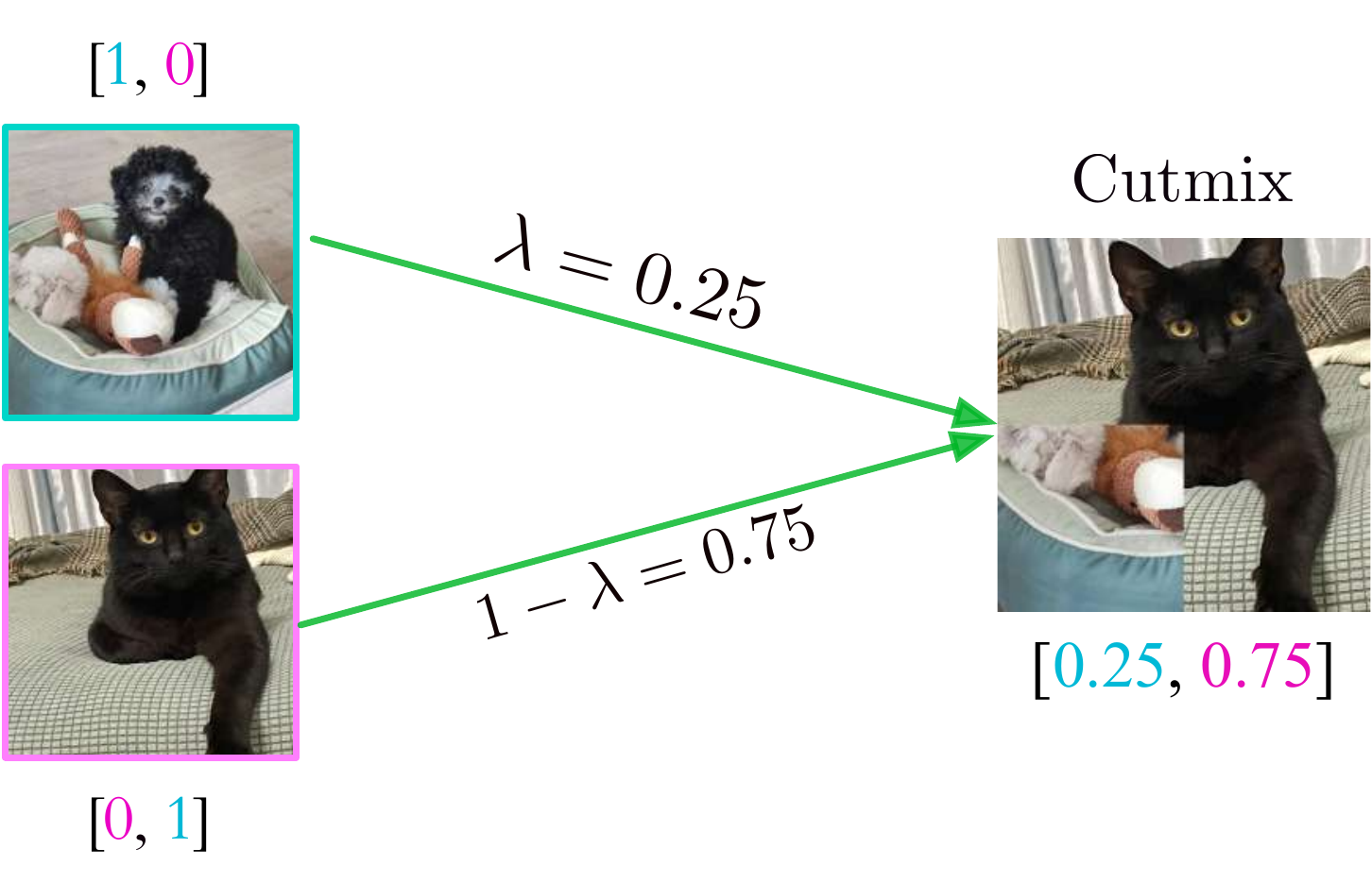}
\caption{Illustration of why Cutmix is problematic when the cut-and-paste patch contains no information about the dog. However, based on the label combination policy, the probability of the dog for the new image is non-zero ($0.25$), misleading the learned model.}
\label{Fig_Cutmixerr}
\end{figure}
Cutmix may be problematic since it aimlessly cuts a patch from a source image and pastes it onto the other image while the Cutmix generated label is proportional to the patch area. It is evident that if the cut-and-paste patch is uninformative for the source image, Cutmix would not improve performance but introduce noise to the target image, leading to unstable training due to the biased constructed target. An example is shown in Figure~\ref{Fig_Cutmixerr} in which the mix ratio $\lambda = 0.25$ and the cut patch is on the bottom left, which does not contain any information about the dog. However, the probability value for the dog in the constructed label is $0.25$, forcing the model to assign a softmax score of $0.25$ to the dog, which is undesirable. 

To overcome the abovementioned issue, \citet{walawalkar2020attentive} put forward Attentive Cutmix, which selects the most illustrative regions based on the feature attention map and cuts the most representative parts. Similarly, FocusMix~\cite{kim2020cut} adopts proper sample approaches to cut-and-paste the instructive regions. Motivated by ViTs~\cite{dosovitskiy2020image}, TransMix~\cite{chen2022transmix} is introduced to mix targets of input pair based on the attention maps learned with ViTs and assign larger values for input images with greater attention. However, directly applying Cutmix on ViTs may lead to a token fluctuation phenomenon, i.e., the contributions of input tokens fluctuate as forward propagation, resulting in a different mix ratio in the output tokens. To handle this issue, token-label alignment~\cite{xiao2022token} is used to track the correspondence between the original and mixed tokens to keep the label for each token by reusing the computed attention at each layer.

SaliencyMix~\cite{uddin2021saliencymix} prefers to cut the indicative regions of the source image with the help of the saliency detector and paste it onto the target image. Four accepted saliency detection algorithms are evaluated, and VSFs~\cite{montabone2010human} performs best on both benchmarks. To explore the effect of the combination strategy, five possible schemes (source to target) are investigated: (i) \textit{Salient} to \textit{Corresponding}, (ii) \textit{Salient} to \textit{Salient}, (iii) \textit{Salient} to \textit{Non-Salient}, (iv) \textit{Non-Salient} to \textit{Salient}, and (v) \textit{Non-Salient} to \textit{Non-Salient} in which \textit{(Non-)Salient} denotes the (non-)salient regions and \textit{Corresponding} indicates the same location. Experiment evaluations show that scheme $\{\textbf{iii}\}$ performs best since it can generate a variety of mixed examples compared with scheme $\{\text{i}\}$ and contains more saliency information compared to schemes $\{\text{ii}, \text{iv}, \text{v}\}$.

Similar to \textit{Salient} to \textit{Non-Salient} strategy in SaliencyMix, \citet{kim2020puzzle} put forward Puzzle Mix, which jointly hunts for (i) the optimal mask and (ii) the optimal transport by taking advantage of the saliency information and the underlying statistics of images. There are two main differences between SaliencyMix and Puzzle Mix: (i) the determination of the Non-Salient region in Puzzle Mix is formalized as an optimization problem rather than stochastically selecting non-informative regions in SaliencyMix and (ii) the saliency information in Puzzle Mix is the $\ell_2$ norm of the gradient values across input channels rather than based on some detection models in SaliencyMix. 

\begin{figure}[!t]
\centering
\includegraphics[width=2.8in]{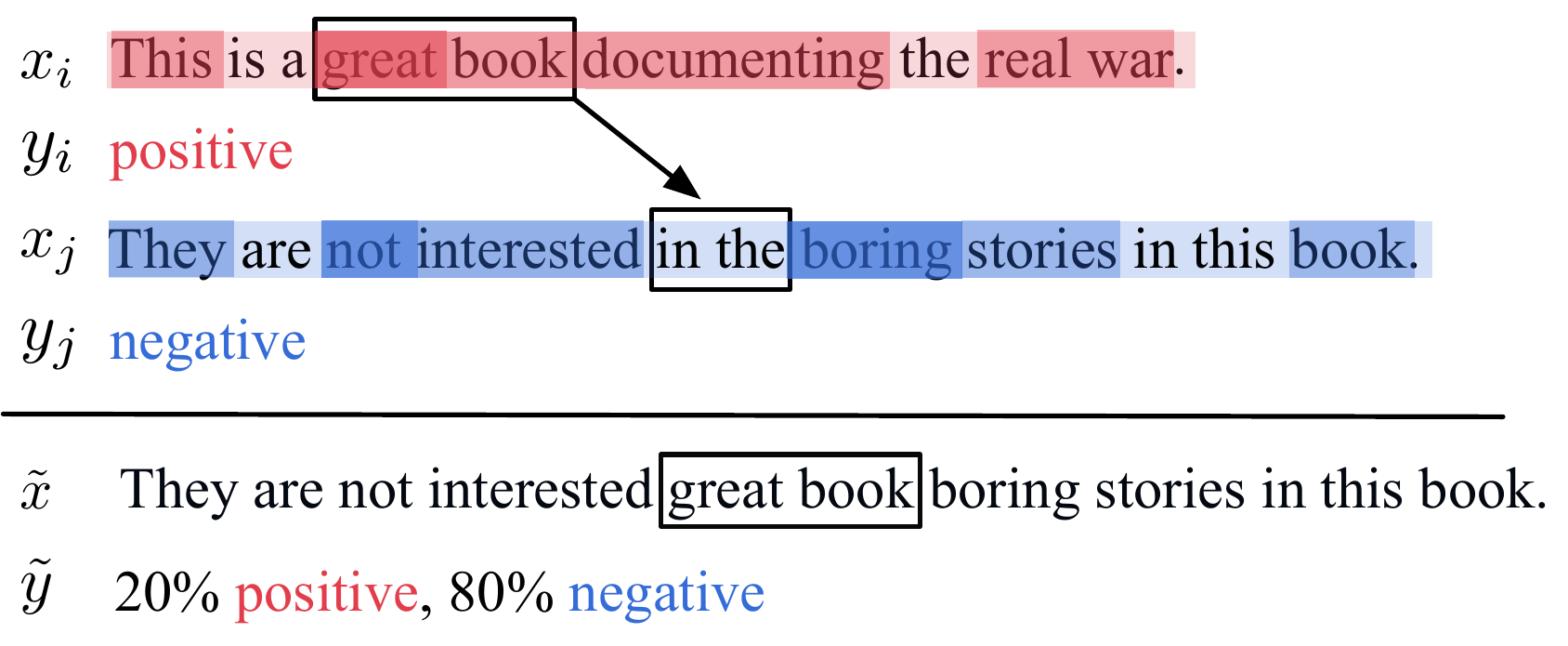}
\caption{Illustration of SSMix. The $j$-th sentence's most irrelevant tokens are replaced with the $i$-th sentence's most related tokens.}
\label{Fig_SSMix}
\end{figure}

To utilize saliency information when mixing two sentences, SSMix~\cite{yoon2021ssmix} produces a new instance while maintaining the locality of the original input through span-based combination and selecting the tokens most related to the prediction based on the magnitude of a gradient. Figure~\ref{Fig_SSMix} illustrates SSMix, where the most $20\%$ ($\lambda$) irrelevant tokens in the $j$-th sentence are replaced with the most $20\%$ related tokens in the $i$-th sentence. Semantically Proportional Mixing (SnapMix)~\cite{huang2021snapmix} calculates the underlying composition of generated images via class activation map (CAM) to reduce label noise. For the fine-grained recognition tasks, \citet{li2020attribute} propose Attribute Mix to exploit semantic information from the input pair at the attribute level. To avoid object information missing and inappropriate mixed labels, ResizeMix~\cite{qin2020resizemix} directly shrinks one image into a small patch and pastes it on a location of the other input stochastically. 

In total, saliency information plays a crucial role in Cutmix-based methods due to its significance in determining the locality and label combination scheme. Without considering saliency information, a random or aimless cut-and-paste process during Cutmix could introduce noise into the constructed features and bias into the generated targets. Therefore, leveraging saliency information can help maximize the benefits of training on Cutmix-ed data.

\begin{table}[ht]
  \centering
  \caption{Summary of selected Cutmix-based Adaptations.}
  \label{Table_CutmixAdaptation}
  \begin{tabular}{r|c|l}
    \hline
    Method & Generated Image & Saliency Map 
    \\ \hline
    SaliencyMix~\cite{uddin2021saliencymix}
    & \begin{minipage}[b]{0.2\columnwidth}
		\centering
		\raisebox{-.5\height}{\includegraphics[width=0.8\linewidth]{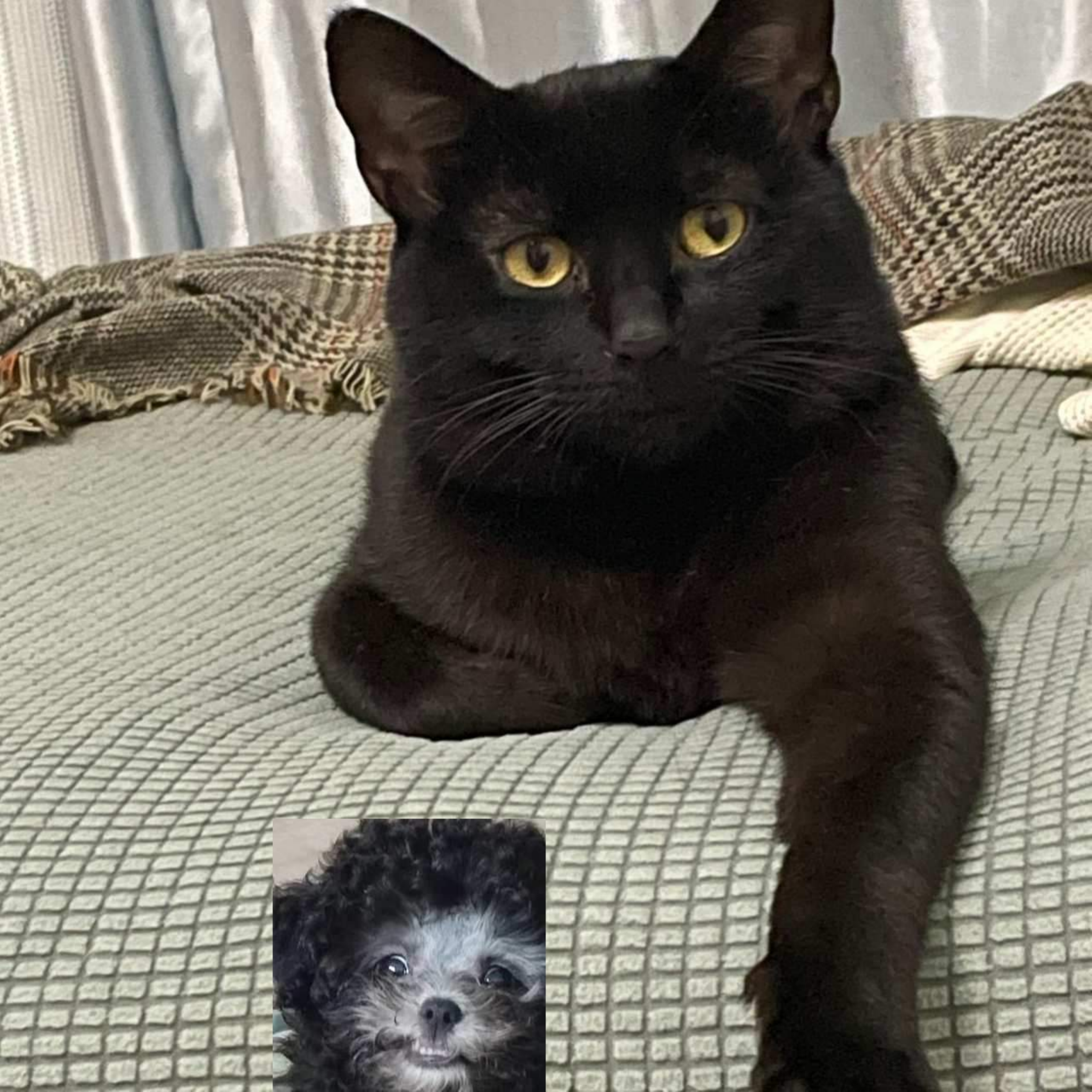}}
	\end{minipage}
    & \begin{minipage}[b]{0.2\columnwidth}
		\centering
		\raisebox{-.5\height}{\includegraphics[width=0.8\linewidth]{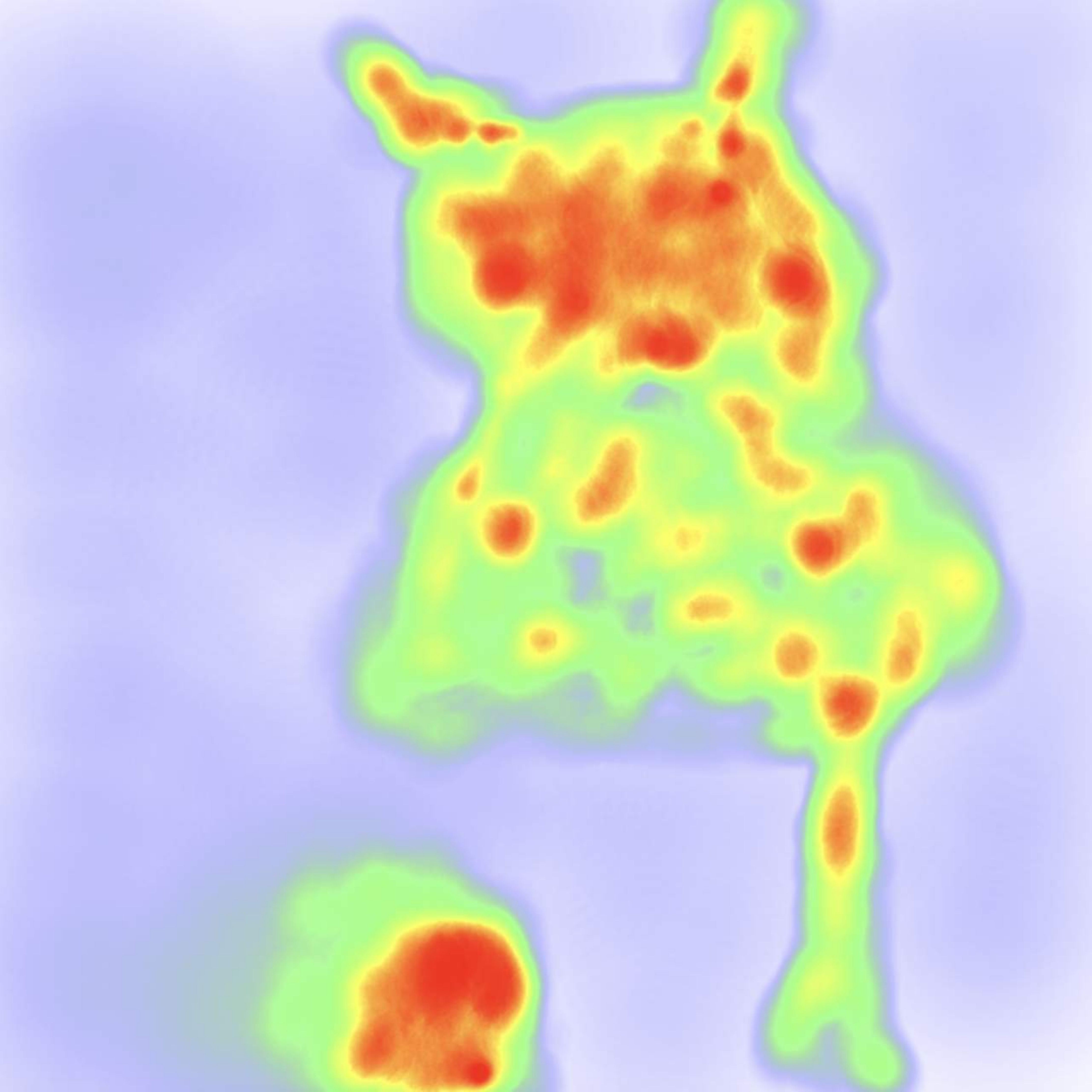}}
	\end{minipage}
    \\ \hline
    Puzzle Mix~\cite{kim2020puzzle}
    & \begin{minipage}[b]{0.2\columnwidth}
		\centering
		\raisebox{-.5\height}{\includegraphics[width=0.8\linewidth]{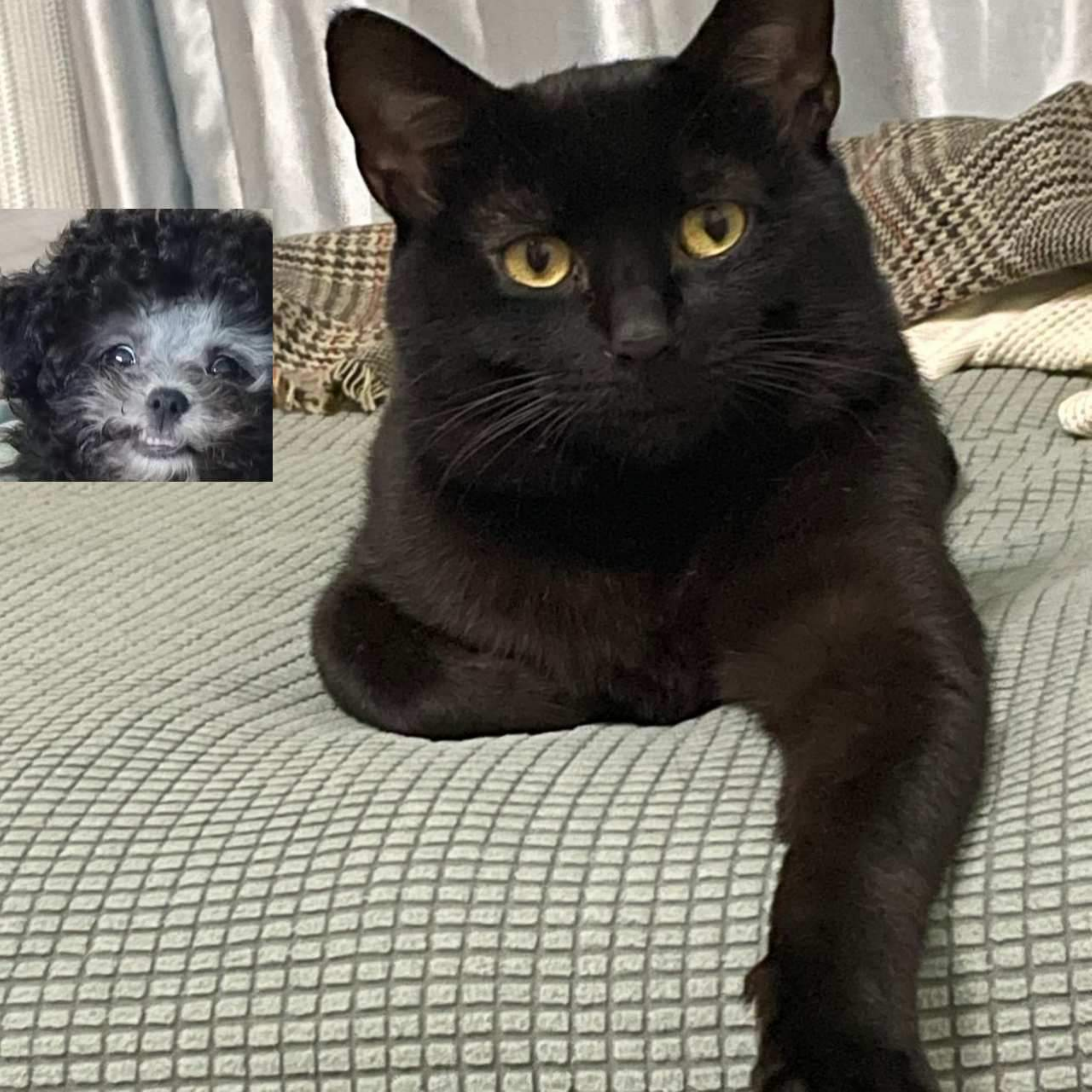}}
	\end{minipage}
    & \begin{minipage}[b]{0.2\columnwidth}
		\centering
		\raisebox{-.5\height}{\includegraphics[width=0.8\linewidth]{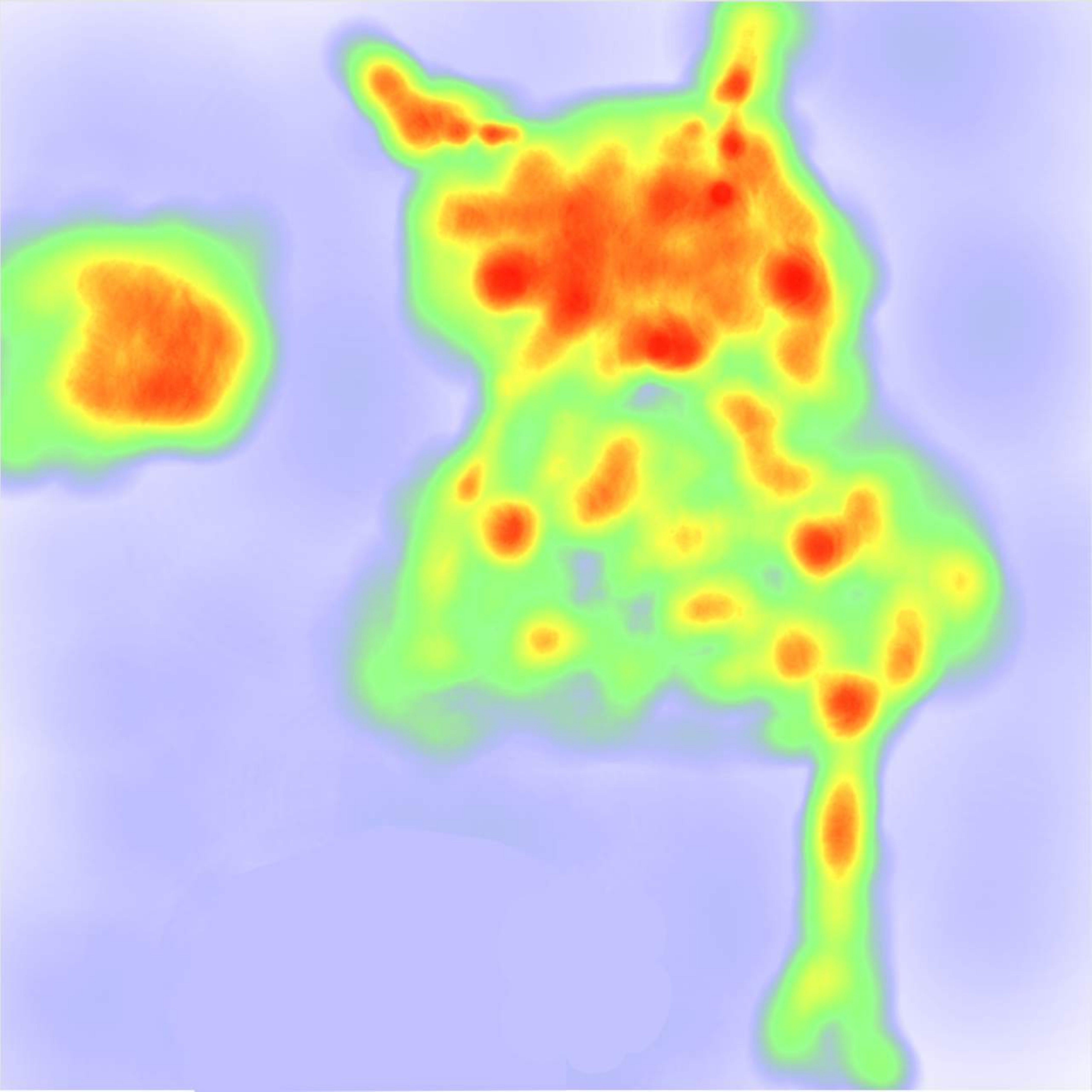}}
	\end{minipage}
    \\ \hline
    ResizeMix~\cite{qin2020resizemix}
    & \begin{minipage}[b]{0.2\columnwidth}
		\centering
		\raisebox{-.5\height}{\includegraphics[width=0.8\linewidth]{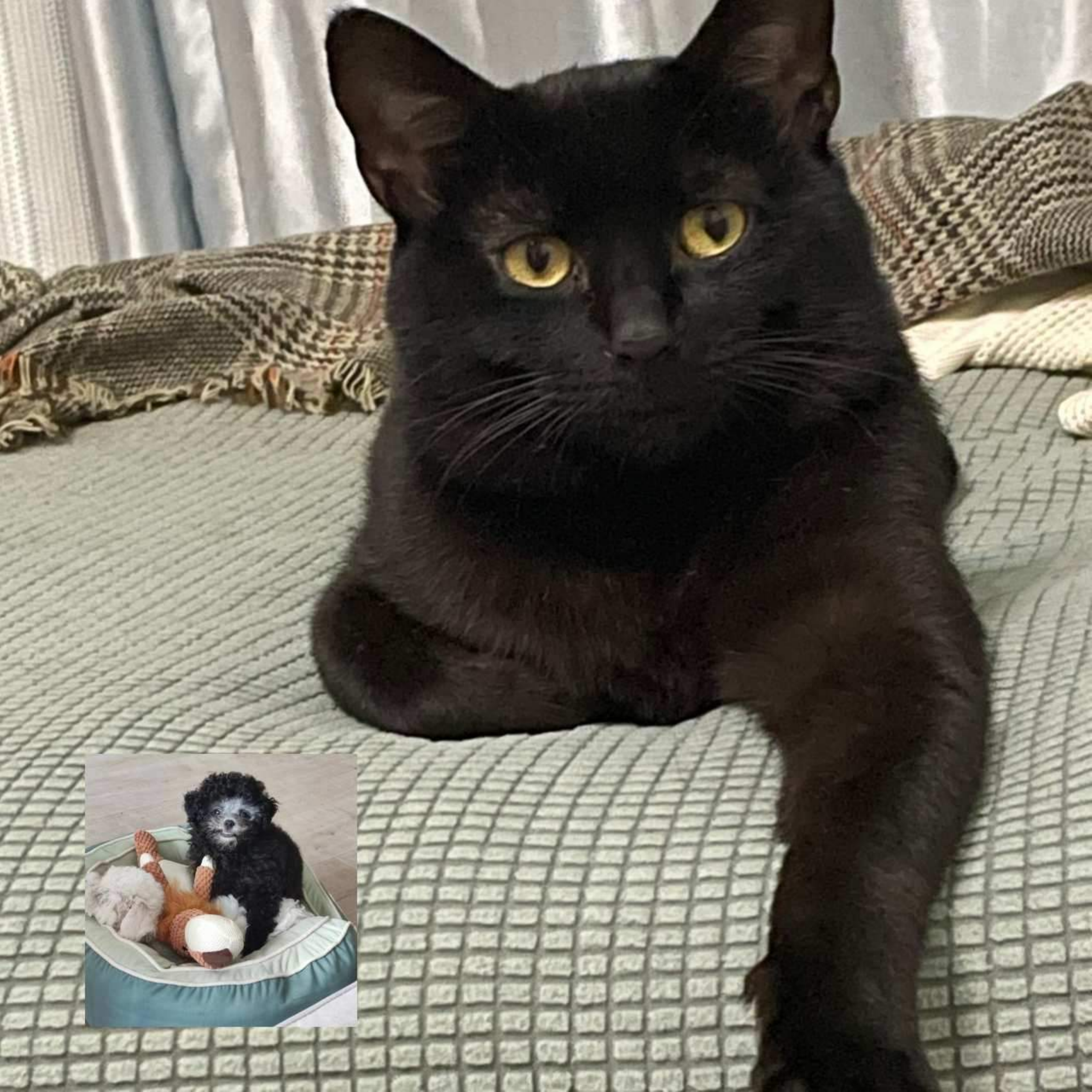}}
	\end{minipage}
    & \begin{minipage}[b]{0.2\columnwidth}
		\centering
		\raisebox{-.5\height}{\includegraphics[width=0.8\linewidth]{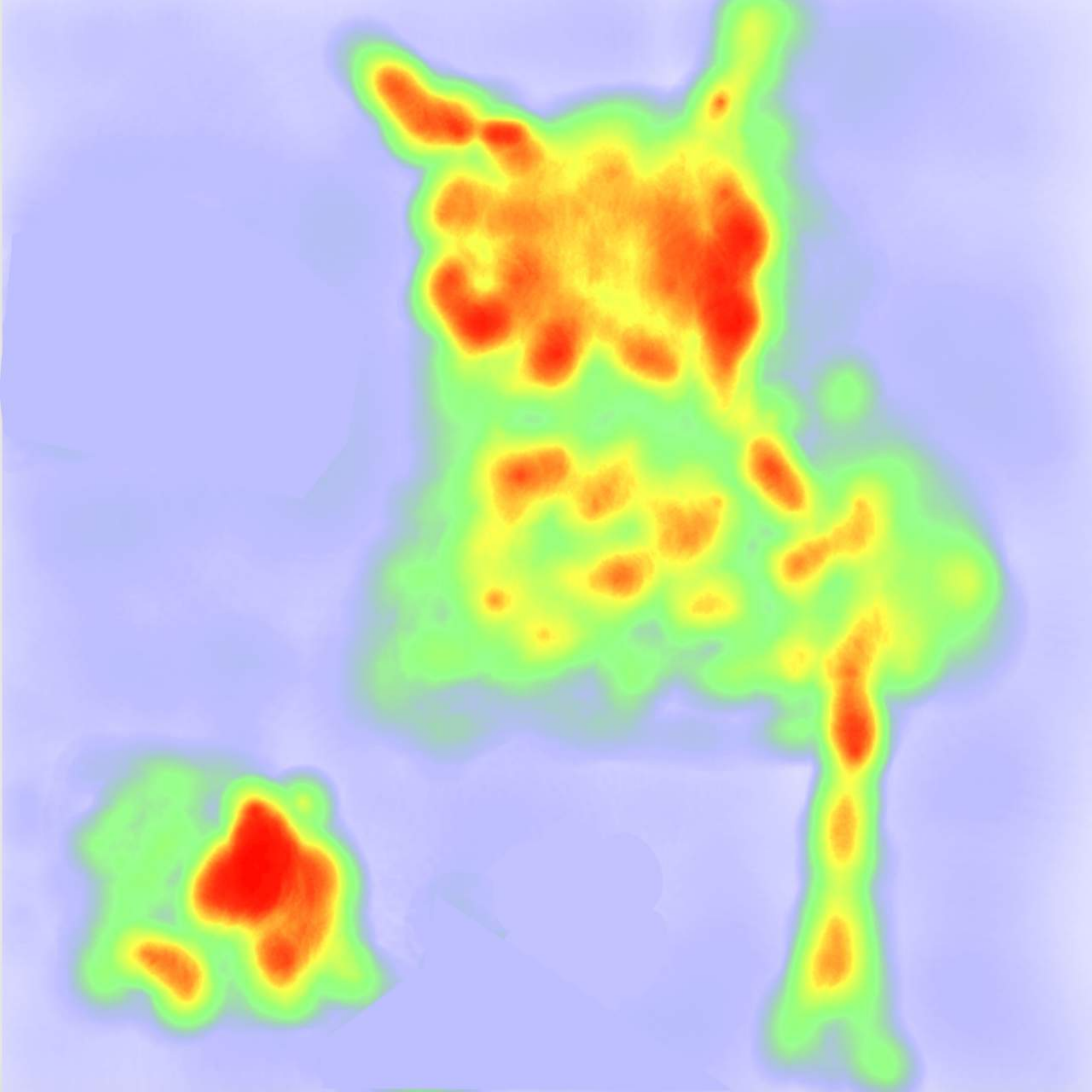}}
	\end{minipage}
    \\ \hline
    Saliency Grafting~\cite{park2022saliency}
    & \begin{minipage}[b]{0.2\columnwidth}
		\centering
		\raisebox{-.5\height}{\includegraphics[width=0.8\linewidth]{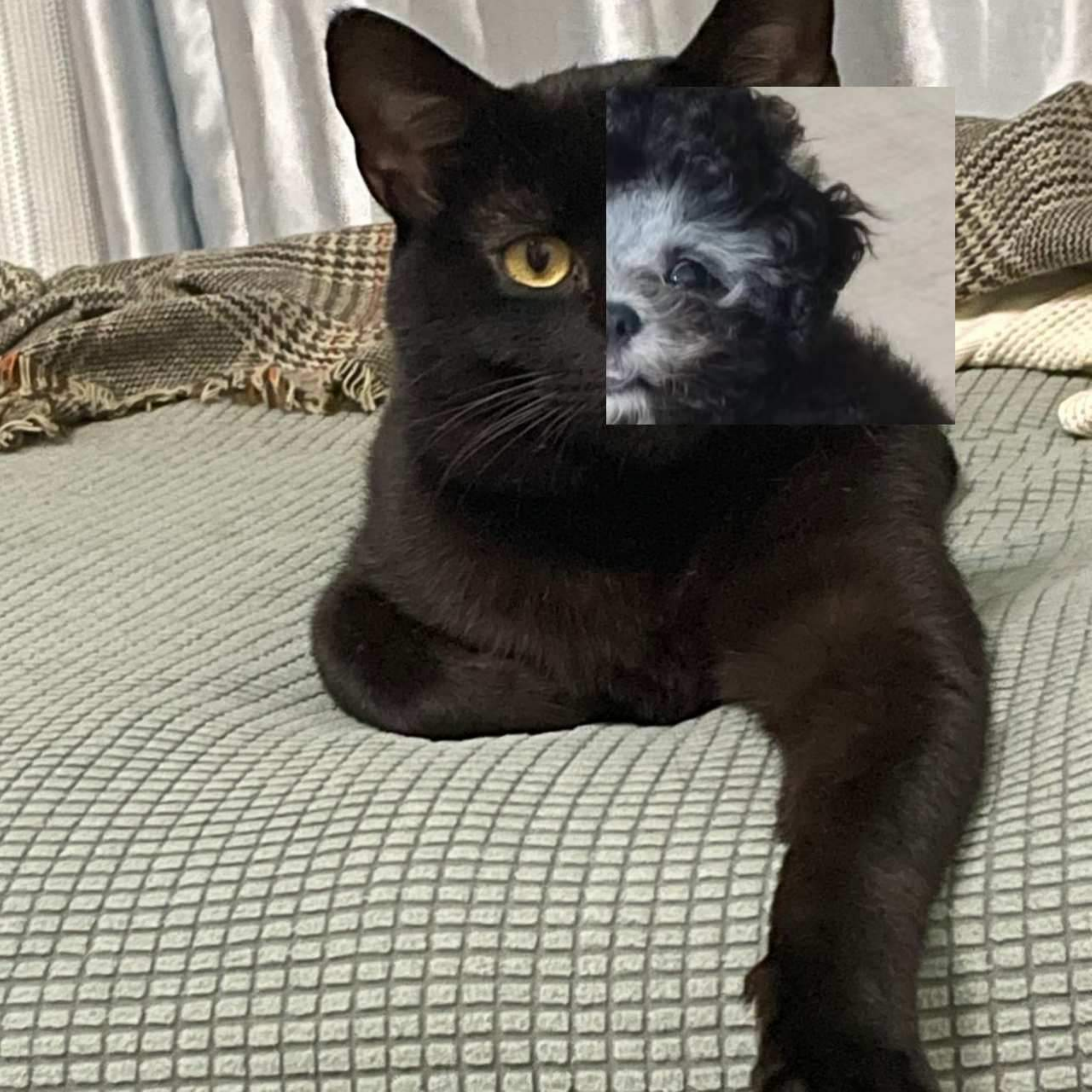}}
	\end{minipage}
    & \begin{minipage}[b]{0.2\columnwidth}
		\centering
		\raisebox{-.5\height}{\includegraphics[width=0.8\linewidth]{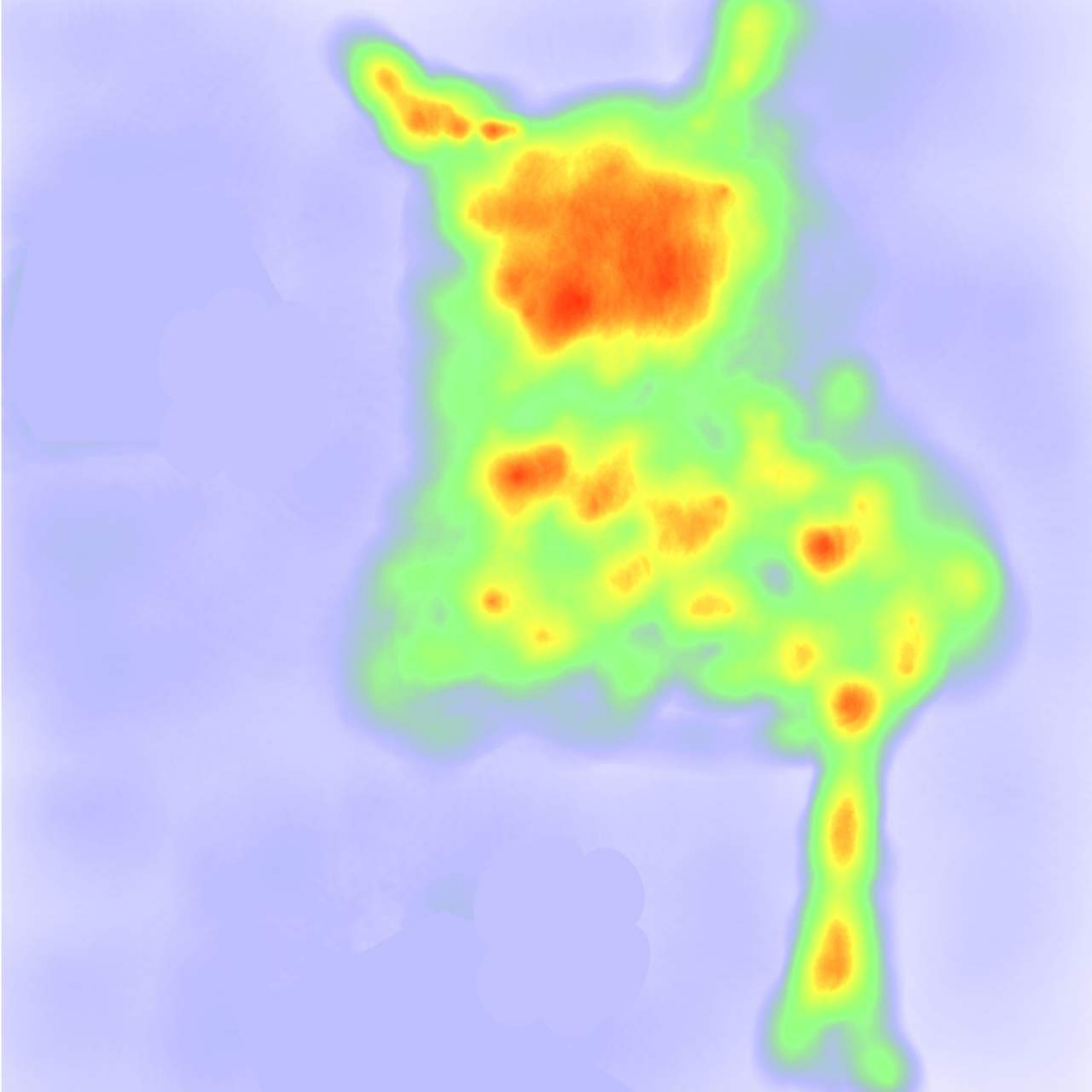}}
	\end{minipage}
    \\ \hline
  \end{tabular}
\end{table} 

\subsubsection{Improved Divergence}
\label{Sub_Divergence}
While saliency-guided methods in Cutmix facilitate preserving essential features and structures, there is a potential drawback related to the diversity of the augmented data. Saliency-guided methods often prioritize specific regions deemed informative. This prioritization results in a concentration of augmentation effects on these selected areas, reducing the diversity of the augmented data. Other regions or less salient features may not receive equal augmentation or attention, which could result in a biased representation of the data distribution.

To get a better trade-off between \textit{plausibility} and \textit{diversity}, Saliency Grafting~\cite{park2022saliency} is designed to synthesize diverse and reasonable examples. Instead of choosing the most informative patch, Saliency Grafting scales and truncates the saliency map to increase the number of options and diversity. Additionally, a Bernoulli distribution is used to sample these candidate regions. Saliency is also utilized to guide the target combination and guarantee the rationality of virtual input-output pairs. Similarly, Co-Mixup~\cite{kim2021co} generates mixed data by maximizing the saliency measure of each mixed sample and the supermodular diversity among the augmented data. Representative Cutmix-based adaptations are summarized in Table~\ref{Table_CutmixAdaptation}, where the generated images with their saliency map are instantiated. To integrate with the accumulated knowledge from the previous iterations, RecursiveMix~\cite{yang2022recursivemix} proposes a recursive mix policy that leverages the historical \textit{Input}-\textit{Prediction}-\textit{Output} triplets. Specifically, RecursiveMix utilizes mixed input-output pairs from the last iteration to generate new synthetic examples, increasing data diversity and encouraging the model to learn from multi-scale and multi-space views. Furthermore, a consistency loss for matching spatial semantics between newly generated samples and the last historical inputs is designed to learn scale-invariant and space-invariant representation.

In general, training with plausibility data can improve performance on in-distribution data while increasing the diversity of the created data, which can enhance the model's robustness to OOD data and adversarial attacks.

\subsubsection{Border Smooth}
Another problem of Cutmix is the "strong-edge" issue, which refers to the sudden pixel change around the bounding box (i.e., the rectangle's border). For example, part of the object in the source images could be cut and then pasted on the region of the target image where the object is located. This issue can result in a noticeable boundary or inconsistency in the augmented image.

To solve this problem, the construction of mask $\mathbf{M}$ in SmoothMix~\cite{lee2020smoothmix} relies on some properties such as width, height, and spread of the input feature map. More specifically, $\mathbf{M}$ is defined as the combination of its center point coordinates $[m, n]$ and spread $\zeta$ in the image space, where $m$ and $n$ are randomly sampled from two uniform distributions with the range of width $W$ and height $H$. $\zeta$ denotes the area of the cut patch, and the smoothing region is also from a uniform distribution. Take the circle mask as an example:
\begin{equation}
\label{Eq_SmoothMix}
    \mathbf{M}_{w} = e^{-\frac{(r-m)^2}{2 \zeta^2}}, \ \mathbf{M}_{h}=e^{-\frac{(s-n)^2}{2 \zeta^2}}, \ \mathbf{M} = \mathbf{M}_{w} \otimes \mathbf{M}_{h},
\end{equation}
where $\otimes$ is the outer product and $r / s$ is the pixel in the width / height dimension. In this way, two inputs are mixed with a smooth transition due to the border gradually diminishing outward. \citet{park2022unified} present a Hybrid strategy of the Mixup and Cutmix (HMix) and a Gaussian Mixup (GMix) scheme. HMix cuts and pastes two samples as Cutmix and then linearly interpolates two images in the area out of the cropped box of Cutmix. In contrast, GMix randomly selects a point and then blends two instances gradually. It uses the Gaussian distribution to relax the Cutmix box condition to a continuous transition, preventing the rectangle cropping of Cutmix from generating implausible synthetic data.

\subsubsection{Other Cutmix Techniques}
Apart from the approaches mentioned above, there are other methods to deal with different problems in Cutmix. Similar to Manifold Mixup~\cite{verma2019manifold}, PatchUp~\cite{faramarzi2020patchup} carries out interpolation in \textit{feature space} -- selecting adjacent regions of feature maps for mixing two examples. The benefit of Cutmix-based techniques on \textit{transformer}-based methods~\cite{vaswani2017attention} is limited since the transformer inherently has a global receptive field, which forces the model to learn from a global perspective. To cope with this issue, \citet{liu2022tokenmix} put forward TokenMix in which the mix of two images is token-level by dividing the input into multiple smaller sub-regions and assigning labels of combined images with CAM from a pre-trained teacher model. A similar work is ScoreMix~\cite{stegmuller2022scorenet}, where the semantic information of images is based on learned self-attention. GridMix~\cite{baek2021gridmix} splits an image into $q \times q$ \textit{grids} and then constructs instances by randomly selecting a local cell from one of the two inputs. The mixed label is determined by blending the label pair in proportion to the number of grids. Besides, an additional task predicting the label of each cell is integrated to extract discriminative representations. One step further, PatchMix~\cite{cascante2021evolving} presents a grid-level mix data augmentation method with a genetic search scheme to find the optimal patch mask. To exploit the spatial context information in remote sensing semantic segmentation data, ChessMix~\cite{pereira2021chessmix} combines transformed mini-patches in a chessboard-like grid to synthesize examples and assigns patches with more examples of the rarest classes the high weight to deal with the imbalance issue. For \textit{unbalanced} classification, Intra-Class Cutmix~\cite{zhao2021intra} boosts model performance by blending intra-class samples of minority classes to modify the decision boundary.

\begin{table*}[ht]
    \caption{Summary of Cutmix-based methods}
    \label{Summary_Cutmix}
    \centering
    \begin{tabular}{r|c|l}
    \hline
    \textbf{Adaptation} & \textbf{Method} & \textbf{Characteristic} \\
    \hline
    \multirow{9}{*}{Integration with Saliency Information}
    & Attentive Cutmix~\cite{walawalkar2020attentive} & Select illustrative regions using feature attention map \\
    & FocusMix~\cite{kim2020cut} & Sample informative pixels  \\
    & TransMix~\cite{chen2022transmix} &  Mix targets using attention maps from ViTs \\
    & SaliencyMix~\cite{uddin2021saliencymix} & Cut indicative regions using saliency detector  \\
    & Puzzle Mix~\cite{kim2020puzzle} & Find optimal mask and transport \\
    & SSMix~\cite{yoon2021ssmix} & Maintain input locality via span-based combination \\
    & SnapMix~\cite{huang2021snapmix} & Compose images using CAM to reduce label noise \\
    & Attribute Mix~\cite{li2020attribute} & Use attribute-level semantic information \\
    & ResizeMix~\cite{qin2020resizemix} & Shrink one input and paste on the other \\
    \hline
    \multirow{3}{*}{Improved Divergence}
    & Saliency Grafting~\cite{park2022saliency} & Scale saliency map and sample regions using Bernoulli \\
    & Co-Mixup~\cite{kim2021co} & Maximize saliency and supermodular diversity  \\
    & RecursiveMix~\cite{yang2022recursivemix} & Use historical Input-Prediction-Output triplets \\
    \hline
    \multirow{3}{*}{Border Smooth}
    & SmoothMix~\cite{lee2020smoothmix} & Construct mask based on image properties \\
    & HMix~\cite{park2022unified} & Cutmix and then Mixup outside cropped box \\
    & GMix~\cite{park2022unified} & Use Gaussian to relax Cutmix to continuous transition \\
    \hline
    \multirow{7}{*}{Others}
    & PatchUp~\cite{faramarzi2020patchup} & Mix intermediate hidden representations \\
    & TokenMix~\cite{liu2022tokenmix} & Mix images at token-level \\
    & ScoreMix~\cite{stegmuller2022scorenet} &  Mix images using learned self-attention semantics \\
    & GridMix~\cite{baek2021gridmix} & Split image into grids and Cutmix cells \\
    & PatchMix~\cite{cascante2021evolving} & Genetic search for optimal patch mask \\
    & ChessMix~\cite{pereira2021chessmix} & Combine mini-patches in chessboard-like grid \\
    & Intra-Class Cutmix & Mix intra-class minority samples \\
    \hline
    \end{tabular}
\end{table*}

\subsection{Beyond Mixup \& Cutmix}
\label{Sub_Others}
Apart from Mixup-based and Cutmix-based methods, there are other techniques based on the mixing principle. These include mixing with itself, incorporating multiple MixDA approaches, and integrating with other DA methods.

\noindent \textbf{Mixing with itself.} DJMix~\cite{hataya2022djmix} combines each training example with its discretized one instead of blending two samples, i.e., mixing only one image. Discretization is conducted by the Vector-Quantized Variational AutoEncoder with encoder $\mathcal{G}(\cdot)$ in an unsupervised manner:
\begin{equation}
\label{Eq_DJMix}
\tilde{\mathbf{x}} = \lambda \mathbf{x} + (1 - \lambda)\mathcal{G}(\mathbf{x}).
\end{equation}
A Jensen–Shannon (JS) divergence-based loss is introduced to map $\mathbf{x}$ and $\tilde{\mathbf{x}}$ closely and obtain consistent representations. CutBlur~\cite{yoo2020rethinking} mixes two resolution versions of an image by cutting and pasting a high-resolution patch to the corresponding low-resolution image area and vice versa. It enforces the model to learn "how" and "where" to super-resolve an image and understand the degree of resolution. 

\noindent \textbf{Incorporating multiple MixDA approaches.} RandomMix~\cite{liu2022randommix} conducts random permutation for data batch to generate two times data and combine multiple mix policies: Mixup~\cite{zhang2018mixup}, Cutmix~\cite{yun2019cutmix}, Fmix~\cite{harris2020fmix}, and ResizeMix~\cite{qin2020resizemix}. Besides, AugRmixAT~\cite{liu2022augrmixat} generated multiple different sets of augmented data for a single example by combining multiple MixDA~\cite{zhang2018mixup, yun2019cutmix, harris2020fmix, qin2020resizemix} and adversarial perturbation~\cite{szegedy2014intriguing} to boost generalization and robustness.

\noindent \textbf{Integrating with other DA methods.} Through the lens of irregular superpixel decomposition, Hammoudi et al. propose SuperpixelGridMix~\cite{hammoudi2022superpixelgridcut}, a new style of data augmentation method that can be combined with mix-based approaches to form their variants. AugMix~\cite{hendrycks2020augmix} presents a data augmentation pipeline with operations from AutoAugment~\cite{cubuk2019autoaugment}. AugMix consists of $3$ sub-chains, each of which is a randomly selected conventional SsDA maneuver. Mixup then combines the generated images with a mix ratio drawn from a Dirichlet distribution $\operatorname{Dirichlet}(\alpha, \alpha, \alpha)$. "skip connection" with Mixup is used to blend the images from sub-chains and the original image. Besides, a JS divergence-based loss function is designed to provide a consistent representation. StackMix~\cite{chen2022stackmix} is an orthogonal work that takes input as the concatenation of two images and the target as the average of two one-hot vectors. It can be combined with mix-based methods. For example, Mixup with StackMix can generate new data based on four inputs, enlarging the space of augmented data compared to the original Mixup that cannot benefit from blending more than two inputs~\cite{zhang2018mixup}.

\subsection{Discussion}
\label{Sub_Summary}
The primary distinction between Mixup and Cutmix lies in their methodologies for combining samples: Mixup integrates samples globally, whereas Cutmix does so locally. Due to its global blending strategy, Mixup is more suited for tasks demanding a comprehensive grasp of the entire image, such as image classification. In contrast, Cutmix, with its localized approach, excels in tasks like object localization, where detailed attention to specific image regions is crucial. Moreover, when the patch used in Cutmix is not informative relative to the source image, it may fail to enhance performance and introduce noise, destabilizing the training process by creating a biased target. Empirical evidence from experiments using ResNet-50 on CIFAR-100 shows that Mixup achieves a top-1 accuracy of $82.10\%$, slightly outperforming Cutmix's $81.67\%$. On the CUB200-2011 benchmarks, however, Cutmix surpasses Mixup in localization accuracy, achieving $54.81\%$ compared to Mixup's $49.30\%$. This is likely because the global approach of Mixup leads to more significant ambiguity in sample generation compared to Cutmix, prompting models trained with Mixup to concentrate on smaller, more discriminating regions of objects. While beneficial for classification, this focus reduces performance in tasks requiring precise localization.

In our detailed analysis of Mixup and Cutmix adaptations, we assess their performance enhancements by integrating them with ResNet-50 and testing on the CIFAR-100 benchmark, as summarized in Table~\ref{Benchmark_Comp2}. Notably, each adaptation offers distinct advantages: the "Adaptive Mix Strategy" most significantly improves Mixup by addressing its inherent challenge of a random mixing ratio. Additionally, enhancements like "Diversity" and "Mixing in Embedding Space" are particularly effective, as they aim to augment data variability and leverage more discriminative embeddings, respectively. Conversely, Cutmix benefits more from adaptations such as "Integration with Saliency Information" and "Border Smooth," which mitigate the abrupt transitions inherent to its patch-based approach. However, it is essential to recognize that each modification introduces its challenges. For instance, "Mixing in Embedding Space" tends to slow convergence due to the complexity of selecting appropriate embedding layers for mixing. Similarly, increasing "Diversity in Mixup" and "Improved Divergence" can cause inconsistencies across different data batches. Moreover, adaptations like "Saliency \& Style Guidance" and "Integration with Saliency Information" necessitate meticulous fine-tuning of hyper-parameters, adding complexity to the data augmentation process.

\begin{table*}[ht]
    \caption{Comparisons within Mixup-based and Cutmix-based data augmentation methods. All results are the top-1 accuracies on the CIFAR-100~\cite{krizhevsky2009learning} benchmark with ResNet-50~\cite{he2016deep}.}
    \label{Benchmark_Comp2}
    \centering
    \begin{tabular}{r|c|c|l}
    \hline
    \textbf{Category} & \textbf{Adaptation} & \textbf{Methods} & \textbf{Performance} \\
    \hline
    \multirow{10}{*}{Mixup} 
    & \multirow{2}{*}{Mixing in Embedding Space}
    & Manifold Mixup~\cite{verma2019manifold} & $83.23\%$ \\
    & & NFM~\cite{lim2022noisy} & $83.74\%$ \\
    \cline{2-4}
    & \multirow{2}{*}{Adaptive Mix Strategy}
    & AdaMixUp~\cite{guo2019mixup} & $84.48\%$ \\
    & & AutoMix~\cite{liu2021unveiling} & $84.35\%$ \\
    \cline{2-4}
    & \multirow{2}{*}{Sample Selection}
    & DMix~\cite{li2021boosting} & $83.07\%$ \\
    & & GenLabel~\cite{sohn2022genlabel} & $83.35\%$ \\
    \cline{2-4}
    & \multirow{2}{*}{Saliency \& Style Guidance}
    & StyleMix~\cite{hong2021stylemix} & $82.34\%$ \\
    & & Mixstyle~\cite{zhou2021domain} & $82.69\%$ \\
    \cline{2-4}
    & \multirow{2}{*}{Diversity}
    & MixMo~\cite{rame2021mixmo} & $83.67\%$ \\
    & & PixMix~\cite{hendrycks2022pixmix} & $83.64\%$ \\
    \hline
    \multirow{6}{*}{Cutmix}
    & \multirow{2}{*}{Integration with Saliency Information}
    & SaliencyMix~\cite{uddin2021saliencymix} & $83.66\%$  \\
    & & Puzzle Mix~\cite{kim2020puzzle} & $83.42\%$  \\
    \cline{2-4}
    & \multirow{2}{*}{Improved Divergence}
    & Co-Mixup~\cite{kim2021co} & $82.47\%$ \\
    & & RecursiveMix~\cite{yang2022recursivemix} & $83.01\%$ \\
    \cline{2-4}
    & \multirow{2}{*}{Border Smooth}
    & SmoothMix~\cite{lee2020smoothmix} & $83.87\%$ \\
    & & HMix~\cite{park2022unified} & $84.02\%$ \\
    \hline
    \end{tabular}
\end{table*}

\section{MixDA Applications}
\label{Sec_Application}
In this section, we will review extensive applications of MixDA.

\subsection{Semi-Supervised Learning}
\label{Sub_SSL}
Semi-supervised learning (SSL) has made significant strides in various domains where unlabeled data is abundant while labeling such data remains challenging due to the need for human resources and expertise.

To integrate MixDA into SSL, \citet{verma2019interpolation} propose Interpolation Consistency Training (ICT) by transiting the decision boundary to low-density regions of data distribution. Specifically, ICT forces the prediction of the interpolation of unlabeled samples to be consistent with the interpolation of the model predictions:
\begin{equation}
\label{Eq_ICT}
    f(\lambda \mathbf{u}_{i} + (1-\lambda)\mathbf{u}_{j}) = \lambda f(\mathbf{u}_{i}) + (1-\lambda)f(\mathbf{u}_{j}),
\end{equation}
where $\mathbf{u}_{i}$ and $\mathbf{u}_{j}$ are sampled unlabeled examples, and $f(\mathbf{u}_{i})$ and $f(\mathbf{u}_{j})$ are the corresponding model predictions. ICT regards the prediction of an unlabeled instance as its target (i.e., guessed label). Similarly, \citet{olsson2021classmix} present ClassMix, which combines unlabeled data to construct new examples. The mixed image is generated by cutting half of one image and pasting it on the other. Its corresponding target is computed based on the network's semantic prediction. 

Another independent work is MixMatch~\cite{berthelot2019mixmatch}, which generates synthetic examples using a batch of labeled samples and another batch of unlabeled instances with guessed labels. Different from ICT, which can be regarded as a case of MixMatch where only unlabeled data is mixed, Mixup is utilized to generate mixed labeled samples and mixed unlabeled examples in MixMatch. One step further, ReMixMatch~\cite{berthelot2019remixmatch} improves the semi-supervised learning performance of MixMatch by distribution alignment and augmentation anchoring. To avoid confirmation bias (i.e., error accumulation) of self-training in MixMatch, DivideMix~\cite{li2020dividemix} rejects the sample's label with high noise and regards it as unlabeled data to improve generalization. Specifically, DivideMix trains two individual Gaussian mixture models (GMMs) to separate the training data into a label set and an unlabeled collection, which will be used to train the main network.

\subsection{Contrastive Learning}
Contrastive Learning (CL) has emerged as a prominent approach within self-supervised learning, focusing on the principle of learning discriminative features by contrasting positive and negative samples. According to \citet{liu2021self}, the core idea is to bring closer the representations of augmented versions (positive examples) of the same data point (anchor sample) and simultaneously drive apart the embeddings of different data points (negative samples). CL's operational mechanism can be illustrated by formulating the InfoNCE contrastive loss. For a given anchor sample $\mathbf{x}_{i}$, its augmented version $\mathbf{x}^{\prime}_{i}$ serves as the positive sample, the set of embeddings from negative samples, denoted as $\mathcal{Q}_{i}$, the InfoNCE loss $\mathcal{L}_{\text{InfoNCE}}$ is mathematically defined as:
\begin{equation}
\begin{gathered}
    \mathbf{h}_{i} = \mathcal{G}(\mathbf{x}_{i}), \ \mathbf{h}^{\prime}_{i} = \mathcal{G}(\mathbf{x}^{\prime}_{i}), \\
    \ \mathbf{h}_{k} = \mathcal{G}(\mathbf{x}_{k}) \ (\mathbf{x}_{k} \in \mathcal{Q}_{i}) \\
    \mathcal{L}_{\text {InfoNCE}}= - \log \frac{e^{(\mathbf{h}_{i} \mathbf{h}^{\prime}_{i} / \tau)}}{\sum_{\mathbf{h}_{k}} e^{(\mathbf{h}_{i} \mathbf{h}_{k} / \tau)}},
\end{gathered}
\end{equation}
where $\tau$ is the temperature factor. The minimization of the InfoNCE loss is central to learning useful representations. By maximizing the similarity between an anchor and its positive sample while minimizing the similarity between the anchor and the negative samples, the model learns to embed similar items closer together in the feature space, even without explicit labels. A straightforward MixDA-based method is Mixup Contrast (MixCo)~\cite{kim2020mixco}, which introduces~\textit{semi-positive} samples that are decoded from the mixture of positive and negative instances to learn fine-grained similarity between representations. The semi-positive sample $\tilde{\mathbf{x}}$ is generated by MixDA:
\begin{equation}
    \tilde{\mathbf{x}} = \lambda \mathbf{x}_{i} + (1 - \lambda) \mathbf{x}_{k}, \ \tilde{\mathbf{h}} = \mathcal{G}(\tilde{\mathbf{x}}),
\end{equation}
where $\mathbf{x}_{k}$ is from the negative sample set $\mathcal{Q}_{i}$. The mix $\tilde{\mathbf{x}}$ is the semi-positive sample for both $\mathbf{x}_{i}$ and $\mathbf{x}_{k}$, with similarity as $\lambda$ and $(1 - \lambda)$, respectively. Therefore, another loss function is derived:
\begin{equation}
    \mathcal{L}_{\text {MixCo}}= - [ \lambda \log \frac{e^{(\tilde{\mathbf{h}} \mathbf{h}^{\prime}_{i} / \tau)}}{\sum_{\mathbf{h}_{k}} e^{(\tilde{\mathbf{h}} \mathbf{h}_{k} / \tau)}} + (1 - \lambda) \log \frac{e^{(\tilde{\mathbf{h}} \mathbf{h}^{\prime}_{i} / \tau)}}{\sum_{\mathbf{h}_{k}} e^{(\tilde{\mathbf{h}} \mathbf{h}_{k} / \tau)}}].
\end{equation}
By combining loss $\mathcal{L}_{\text {InfoNCE}}$ and loss $\mathcal{L}_{\text {MixCo}}$, the model can capture such a semi-positive relation, which is much more complicated than merely discriminating positives from the negatives, leading to more efficient use of negatives. \citet{zhang2021unleashing} demonstrate that learning with contrastive loss benefits representation learning and fine-tuning. Contrast-Regularized tuning (Core-tuning) is proposed to fine-tune contrastive self-supervised learning models. A hard pair excavation scheme is applied to learn transferable features and smooth the decision boundary. In particular, Mixup is used to generate (i) \textit{hard positive} pair by mixing the two hardest pairs and (ii) \textit{semi-hard negative} pair by blending a negative example and the anchor data. Other similar works include Feature Transformation (FT)~\cite{zhu2021improving}, mixing of contrastive hard negatives (MoCHi)~\cite{kalantidis2020hard}, and background mixing~\cite{sahoo2021contrast}.

Following the Single Instance Multi-view (SIM) paradigm, Siamese network~\cite{chen2021exploring} takes the augmented images of an example as inputs and conducts an element-wise maximum of features to compute the discriminative representation. The loss function for SIM $\mathcal{L}_{SIM}$ is as follows:
\begin{equation}
\begin{gathered}
        \mathbf{h}_{i} = \mathcal{G}(\mathbf{x}_{i}), \ \mathbf{h}^{\prime}_{i} = \mathcal{G}(\mathbf{x}^{\prime}_{i}), \\
        \mathbf{z}_{i} = \mathcal{P}(\mathbf{h}_{i}), \ \mathbf{z}^{\prime}_{i} = \mathcal{P}(\mathbf{h}^{\prime}_{i}), \\
        \mathcal{D}(\mathbf{h}_{i}, \mathbf{z}^{\prime}_{i})=-\frac{\mathbf{h}_{i}}{\|\mathbf{h}_{i}\|_2} \cdot \frac{\mathbf{z}^{\prime}_{i}}{\|\mathbf{z}^{\prime}_{i}\|_2}, \
        \mathcal{D}(\mathbf{h}^{\prime}_{i}, \mathbf{z}_{i})=-\frac{\mathbf{h}^{\prime}_{i}}{\|\mathbf{h}^{\prime}_{i}\|_2} \cdot \frac{\mathbf{z}_{i}}{\|\mathbf{z}_{i}\|_2}, \\
        \mathcal{L}_{\text{SIM}} = \frac{1}{2} \mathcal{D}(\mathbf{h}_{i}, \mathbf{z}^{\prime}_{i}) + \frac{1}{2} \mathcal{D}(\mathbf{h}^{\prime}_{i}, \mathbf{z}_{i}),
\end{gathered}
\end{equation}
where $\mathcal{P}$ is the prediction head. Guo et al.~\cite{guo2021mixsiam} present MixSiam, a MixDA that feeds the combinations of augmented images into the backbone and encourages the embeddings to be close to the original distinctive representation:
\begin{equation}
\begin{gathered}
        \tilde{\mathbf{x}} = \lambda \mathbf{x}_{i} + (1 - \lambda) \mathbf{x}^{\prime}_{i}, \\
        \mathbf{h}_{i} = \mathcal{G}(\mathbf{x}_{i}), \ \mathbf{h}^{\prime}_{i} = \mathcal{G}(\mathbf{x}^{\prime}_{i}),  \ \tilde{\mathbf{h}} = \mathcal{G}(\tilde{\mathbf{x}}), \\
        \mathbf{z}_{i} = \mathcal{P}(\mathbf{h}_{i}), \ \mathbf{z}^{\prime}_{i} = \mathcal{P}(\mathbf{h}^{\prime}_{i}), \ \tilde{\mathbf{z}} = \operatorname{Max}(\mathbf{z}_{i}, \mathbf{z}^{\prime}_{i}), \\
        \mathcal{L}_{\text{MixSiam}} = \mathcal{D}(\tilde{\mathbf{h}}, \tilde{\mathbf{z}}) = -\frac{\tilde{\mathbf{h}}}{\|\tilde{\mathbf{h}}\|_2} \cdot \frac{\tilde{\mathbf{z}}}{\|\tilde{\mathbf{z}}\|_2},
\end{gathered}
\end{equation}
where $\operatorname{Max}$ denotes the element-wise maximum. By combining $\mathcal{L}_{\text{MixSiam}}$ and $\mathcal{L}_{\text{SIM}}$, MixSiam increases the diversity of the augmented sample and keeps predicted embeddings invariant. However, Siamese frameworks are prone to overfitting if the augmentation technique used to construct two views is not expressive enough. To deal with this problem, unsupervised image mixtures (Un-Mix)~\cite{shen2022mix} entails the soft distance concept into the label space and makes the model embrace the soft degree of similarity between image pairs by interpolating in the input space. Specifically, Un-Mix first generates two views with a pre-defined transformation, followed by a mix operation. Finally, the mixed samples are fed into the two-branch framework to produce hidden representations.

Going beyond SIM learning, BSIM~\cite{chu2020beyond} takes the similarity of spurious-positive pairs (i.e., two randomly sampled instances and their mixture) into account for more even feature distribution, enhancing the discrimination capability of the model. Contrastive learning is recast as learning a non-parametric classifier that assigns a virtual class to each sample in work~\cite{lee2021mix}, which uses $i$-Mix to give a virtual label to each sample and conducts interpolation in both input space and virtual label space to regularize the contrastive representation learning. MCL~\cite{wickstrom2022mixing} designs a novel contrastive loss function to take advantage of the MixDA strategy, which predicts the mixing coefficient regarded as the soft target in the loss function. \citet{li2020center} present CLIM, which finds local examples similar to a cluster center and utilizes a mix scheme to perform smoothing regularization to deduce powerful representation. Simple data mixing prior (SDMP)~\cite{ren2022simple} considers the connections between source images and the generated counterparts and encodes the relationships into hidden space to model this prior. Besides, Similarity Mixup~\cite{patel2022recall} virtually increases the batch size and operates on pair-wise scalar similarities. Many constructed hard negatives in Graph Contrastive Learning (GCL) are positive samples, which will undermine the performance. To resolve this problem, ProGCL-Mix~\cite{xia2022progcl} is proposed to generate more hard negatives considering the probability of a negative being the true one.

Overall, MixDA is utilized to enhance Contrastive Learning (CL) in two main ways: (i) generating diverse mixed examples, such as semi-hard negative pairs and semi-positive samples, to incorporate finer-grained information and facilitate the learning of more discriminative representations and (ii) integrating MixDA into the optimization process as prior knowledge to regularize the CL objective.

\subsection{Adversarial Training}
Adversarial training~\cite{szegedy2014intriguing} can significantly lift model robustness since it encourages the model to explore some unseen regions. This is achieved by perturbing a training data point in the direction of another instance while keeping its origin target as the training label, improving generalization.

Through the lens of optimization, \citet{madry2018towards} formulate adversarial robustness as a min-max problem:
\begin{equation}
    f^{*} = \arg\min_{f \in \mathcal{F}} \int \max_{\delta \in \mathcal{S}} \mathcal{L}(f(\mathbf{x} + \delta), \mathbf{y}) \mathrm{d} P_\psi(\mathbf{x}, \mathbf{y}),
\end{equation}
where $\mathcal{S}$ is the neighborhood region of each sample. $P_\psi(\mathbf{x}, \mathbf{y}) = \frac{1}{N} \sum_{i=1}^{N} \rho(\mathbf{x}=\mathbf{x}_{i}, \mathbf{y}=\mathbf{y}_{i})$ in which $\rho(\mathbf{x}=\mathbf{x}_{i}, \mathbf{y}=\mathbf{y}_{i})$ is a Dirac mass centered at $(\mathbf{x}_{i}, \mathbf{y}_{i})$. It is deliberately designed to have visually imperceptible perturbations $\delta$ only. Adversarial training consists of two procedures: (i) generating adversarial examples and (ii) training the model to assign the origin target to augmented data for improved robustness against unseen adversarial examples. The similarity between adversarial training and MixDA is formulated by reinterpreting MixDA as a kind of Directional Adversarial Training (DAT)~\cite{archambault2019mixup}, which perturbs one training sample based on the direction of the other instance as the same expected loss functions they have. Meanwhile, Empirical Risk Minimization (ERM) is notorious as it can be easily fooled by adversarial examples~\cite{goodfellow2015explaining,szegedy2014intriguing}. In contrast, MixDA can be regarded as Vicinal Risk Minimization (VRM)~\cite{zhang2018mixup}. Besides, from the view of robust objectives and the optimization dynamic for neural networks, adversarial training needs more training data to gain better generalization performance~\cite{schmidt2018adversarially}. Thus, deeply investigating the connection between MixDA and adversarial training would be natural.

By investigating the mutual information between the function learned on the original data and the counterpart on the mixed data by a VAE, Harris et al.~\cite{harris2020fmix} demonstrate that MixDA approaches can be regarded as a form of adversarial training~\cite{goodfellow2015explaining}, therefore, enhancing robustness to attacks such as injecting uniform noise to create examples similar to those synthesized by Mixup. \citet{zhang2021does} attribute the adversarial robustness gain of mix training to the approximate loss function, which acts as an upper bound of the adversarial loss with the $\ell_2$ attack. For robustness, minimizing the loss function with augmented instances is approximately equivalent to minimizing an upper bound of the adversarial loss. This explains why models trained with mixed data manifest robustness improvement for several adversarial attacks. To increase training data, Interpolated Adversarial Training (IAT)~\cite{lamb2019interpolated} trains the model on the mixture of adversarial samples along with the combination of unperturbed instances. Similarly, a soft-labeled approach for improved adversarial generalization, Adversarial Vertex Mixup (AVMixup)~\cite{lee2020adversarial}, is presented to extend the training distribution via the interpolation between the raw input vector and the virtual vector (adversarial vertex) defined in the adversarial direction. \citet{si2021better} put forward Adversarial and Mixup Data Augmentation (AMDA) to explore a much larger attack search space by linearly interpolating the representation pair to generate augmented data. The generated data are more diverse and abundant than discrete text adversarial examples generated by general adversarial methods. Taking advantage of Mixup, M-TLAT~\cite{laugros2020addressing} designs a novel adversarial training algorithm called Targeted Labeling Adversarial Training (TLAT) to boost the robustness of image classifiers for $19$ common corruptions and $5$ adversarial attacks, without sacrificing the accuracy on clean samples. Specifically, M-TLAT interpolates the target labels of adversarial samples with the ground-truth labels:
\begin{equation}
\label{Eq_M-TLAT}
    \mathbf{x}_{\text{mtlat}}=\tilde{\mathbf{x}}- \arg \max_{\delta \in \mathcal{S}}\{\mathcal{L}(\tilde{\mathbf{x}}+\delta, \mathbf{y}_{\text{target}})\}, \ \mathbf{y}_{\text{mtlat}}=(1-\delta) * \tilde{\mathbf{y}} + \delta * \mathbf{y}_{\text{target}},
\end{equation}
where $(\tilde{\mathbf{x}}, \tilde{\mathbf{y}})$ is the Mixup-ed input-output pair and $\mathbf{y}_{\text{target}}$ is the targeted label for adversarial attack. $\mathbf{x}_{\text{mtlat}}$ contains features from three different sources: two clean images specified by $\tilde{\mathbf{x}}$ and an adversarial perturbation targeting a specific class $\mathbf{y}_{\text{target}}$. Correspondingly, label $\mathbf{y}_{\text{mtlat}}$ embraces their supervised signals with the associated weights, which are jointly decided by the mix ratio $\lambda$ and the perturbation $\delta$. M-TLAT predicts the label corresponding to the three sources while outputting the weight of each source within the combined feature $\mathbf{x}_{\text{mtlat}}$, encouraging the model to learn a subtle representation of mixed images. Bunk et al.~\cite{bunk2021adversarially} combine Mixup and adversarial training by mining the space between samples using projected gradient descent. It utilizes back-propagation through the Mixup interpolation in the training stage to optimize for the matte and incongruous region. Besides, the effect of Mixup ratio optimization is also investigated to improve robustness. Similarly, MixACM~\cite{muhammad2021mixacm} aligns activated channel maps (ACM) by transferring the robustness of a teacher to a student. Mixup-SSAT~\cite{jiao2022semi} mixes the adversarial and benign scenarios in an adversarial training manner to gain a trade-off between accuracy and adversarial training. Moreover, an approach called Clean Feature Mixup (CFM) is introduced by~\citet{byun2023introducing} to enhance the transferability of targeted adversarial examples.

The methods mentioned above implicitly defend against adversarial attacks in the testing stage by directly classifying the test data in a vanilla model trained on mixed data. In contrast, an inference principle called Mixup Inference (MI)~\cite{pang2020mixup} exploits the induced global linearity in the trained model. The basic idea is that breaking the locality via the globality of the model predictions would benefit the adversarial robustness. MI mixes one input with some other randomly chosen clean samples. They can be transformed into equivalent perturbations, destroying the locality of adversarial attacks and reducing their strength. Specifically, MI first samples a label $\mathbf{y}_{s}$ and then samples $K$ examples $\{\mathbf{x}_{s,k}\}^{K}_{k=1}$ from the sampled class. In the inference stage, Mixup is used to generate input $\tilde{\mathbf{x}}_{s,k} = \lambda\mathbf{x} + (1-\lambda)\mathbf{x}_{s,k}$ and the final result is averaged on the output predictions of the $K$ Mixup-ed examples:
\begin{equation}
\label{Eq_MI}
    f(\mathbf{x})=f(\mathbf{x}) + \frac{1}{K}f(\tilde{\mathbf{x}}_{s,k}).
\end{equation}

\subsection{Generative Models}
Generative models, such as GANs~\cite{goodfellow2014generative}, VAE~\cite{kingma2014auto}, normalizing flows~\cite{kobyzev2020normalizing}, and denoising diffusion probabilistic models (DDPM)~\cite{ho2020denoising}, learn underlying probability distribution to understand data and generate new instances with new configurations of latent presentation. An autoencoder model $\mathcal{F}(\cdot)$ consists of an encoder $\mathcal{G}(\cdot)$ and a decoder $\mathcal{H}(\cdot)$, i.e., $\mathcal{F}(\cdot)$ = $\mathcal{H}(\mathcal{G}(\cdot))$. An input $\mathbf{x}$ is encoded as a hidden embedding $\mathbf{z}$ via $\mathcal{G}$ and then reconstructed by $\mathcal{H}$. 

Applying MixDA in the generative models is straightforward. For example, pseudo input is generated by Mixup and then leveraged to break the evidence lower bound (ELBO) value bottleneck by reducing the margin between ELBO and the data likelihood~\cite{feng2021shot}. Another two direct applications of MixDA are adversarial autoencoder (AAE)~\cite{liu2018data} and VarMixup~\cite{mangla2020varmixup}. The former interpolates hidden features to construct a broad of latent representations. The latter linearly interpolates on the unfolded latent manifold where the linearity of training data points is preserved.

\citet{berthelot2019understanding} suggest that, in some cases, autoencoders can be interpolable, i.e., decoding the convex combination of the representations in the hidden space for two data points, the autoencoder can generate semantically mixed instances. Accordingly, Adversarially Constrained Autoencoder Interpolation (ACAI) utilizes a critic network $d$ to force interpolated data points to be realistic, and the training objective is to minimize:
\begin{equation}
\label{Eq_ACAI}
    \mathcal{L}_{d} = \|d(\mathcal{H}(\lambda \mathcal{G}(\mathbf{x}_{i} + (1 - \lambda)\mathcal{G}(\mathbf{x}_j)) - \lambda)) \|^{2} + \ \|d(\eta\mathbf{x} + (1 - \eta)\mathcal{F}(\mathbf{x})) \|^{2},
\end{equation} 
where $\eta$ is a scalar hyperparameter. The first term aims to recover the mix ratio used for interpolation in latent space, while the second term is a regularizer with two utilities: (i) it encourages the critic consistently to output $0$ for non-interpolated inputs and (ii) it enables the critic to obtain realistic data by interpolating in data space, i.e., $\mathbf{x}$ and $\mathcal{F}(\mathbf{x})$, even when the autoencoder's reconstructions are poor. 

To improve the quality of reconstructed output, \citet{larsen2016autoencoding} combine it with an adversarial game. In particular, a discriminator $\mathcal{D}$ attempts to discern between real input and reconstructed output. In turn, the autoencoder is expected to generate realistic reconstructions to fool the discriminator, formulating a new baseline -- \text{AE\_GANs}:
\begin{equation}
\label{Eq_AEGAN1}
    \min_{\mathcal{F}} \mathbb{E}_{\mathbf{x} \sim P}\underbrace{\|\mathbf{x}-\mathcal{F}(\mathbf{x})\|_2}_{\text{reconstruction}}+\eta\underbrace{\mathcal{L}_\text{GAN}(\mathcal{D}(\mathcal{F}(\mathbf{x})), 1)}_{\text{fool}\,D\,\text{with reconstruction}}, 
\end{equation}
\begin{equation}
\label{Eq_AEGAN2}
    \min_{\mathcal{F}} \mathbb{E}_{\mathbf{x} \sim P} \underbrace{\mathcal{L}_\text{GAN}(\mathcal{D}(\mathbf{x}), 1)}_{\text{label}\,\mathbf{x}\,{as\,\text{real}}}+\underbrace{\mathcal{L}_\text{GAN}(\mathcal{D}(\mathcal{F}(\mathbf{x})), 0)}_{\text{label reconstruction as fake}},
\end{equation}
where $\mathcal{L}_\text{GAN}$ is the binary cross-entropy equivalent to the JS GANs~\cite{goodfellow2014generative}. Based on this, adversarial Mixup resynthesis (AMR)~\cite{beckham2019adversarial} mixes hidden embeddings that are indistinguishable from real samples after decoding. AMR first encodes a hidden pair $\mathbf{h}_{i} = \mathcal{G}(\mathbf{x}_{i})$ and $\mathbf{h}_{j} = \mathcal{G}(\mathbf{x}_{j})$ and then mixes them. The mixed representation is decoded by the decoder $\mathcal{H}(\cdot)$. AMR minimizes a tailored loss function that aims to fool the discriminator $\mathcal{D}$ with output decoded from the constructed instances. Formally, term $\mathcal{L}_\text{GAN}(\mathcal{D}(\mathcal{H}(\operatorname{Mix}(\mathbf{h}_{i}, \mathbf{h}_{j}))), 1)$ is added to the loss function of autoencoder: Equation~\eqref{Eq_AEGAN1}, and term $\mathcal{L}_\text{GAN}(\mathcal{D}(\mathcal{H}(\operatorname{Mix}(\mathbf{h}_{i}, \mathbf{h}_{j}))), 0)$ is appended to the loss function of GANs: Equation~\eqref{Eq_AEGAN2}.

\subsection{Domain Adaptation}
Domain adaptation involves transferring a model that is trained on a labeled source domain $\mathcal{X}{s}$ to an unlabeled target domain $\mathcal{U}{t}$. Like semi-supervised learning (SSL) discussed earlier, MixDA is also an effective approach for leveraging unlabeled data in domain adaptation. By incorporating MixDA, the model can benefit from the additional information in the unlabeled target domain, leading to improved performance and adaptation to the target domain.

Mao et al.~\cite{mao2019virtual} present a novel general algorithm called Virtual Mixup Training (VMT) to impose the linear combination constraint on the region in-between training data for improved regularization. Unlike conventionally mixing samples with labels, VMT constructs new data by combining examples without any supervision signal from the target domain. To do this, VMT synthesizes virtual samples with guessed labels as:
\begin{equation}
\label{Eq_VMT}
    \tilde{\mathbf{u}} = \lambda \mathbf{u}_{i} + (1-\lambda)\mathbf{u}_{j}, \ \tilde{\mathbf{y}} = \lambda f(\mathbf{u}_{i}) + (1-\lambda)f(\mathbf{u}_{j}), \ \mathcal{L} = \mathbb{E}[\operatorname{D}_\text{KL}(\tilde{\mathbf{y}},f(\tilde{\mathbf{u}}))],
\end{equation}
where $\mathbf{u}_{i}$ and $\mathbf{u}_{j}$ are the samples without labels from the target domain $\mathcal{U}_{t}$ and $\operatorname{D}_\text{KL}$ denotes the KL divergence. However, training VMT with conditional entropy loss may be unstable and may even result in a degenerated solution for some complicated tasks. To tackle this problem, VMT mixes on the input of the softmax layer (i.e., logit values) rather than the output of the softmax layer (i.e., probabilities), as most of the probability values are encouraged to be zero due to the one-hot target vector, while the logits still have non-zero values. Inter- and Intra-domain Mixup training (IIMT)~\cite{yan2020improve} imposes training constraints across domains using Mixup to enhance the generalization performance on the target domain, where adversarial domain learning and Intra-domain Mixup are explored to improve performance further. Another work combining adversarial domain adaptation and Mixup is DM-ADA~\cite{xu2020adversarial}, which keeps domain invariance in a continuous latent space and trains a domain discriminator to discern examples relative to the source and target domains. The mixed domain label is then used to encourage the domain discriminator to align representations and constrain distance between mixed examples. Besides, domain Mixup is jointly implemented on both pixel and feature levels to lift the model's robustness. Similarly, Dual Mixup regularized learning (DMRL)~\cite{wu2020dual} forces the model to output coherent predictions to boost the intrinsic structures of the hidden space by inherent structures carrying domain and category Mixup regularization. To mitigate intra-domain discrepancy and inter-domain representation gap, a progressive data interpolation strategy is proposed by~\citet{ma2023source}, incorporating progressive anchor selection and dynamic interpolation rate. The negative transfer -- the dearth of domain invariance and distinction of the latent representation -- is the main issue in partial domain adaptation, where the label space of the target domain is a subset of the source label space. To deal with this problem, a Select, Label, and Mix (SLM) framework~\cite{sahoo2021select} is developed to discriminate the invariant feature embedding. The Mix conducts domain Mixup with the Select and Label parts to scrutinize more intrinsic structures across domains, resulting in a domain-invariant hidden space.

\subsection{Sentence Classification}
\label{Sub_NLP}
Conventional DA techniques for text data~\cite{feng2021survey,li2022data} differ from those for image data~\cite{yang2022image,shorten2019survey}, which can directly leverage human knowledge for label-invariant data transformation, such as cropping, flipping, and changing the intensity of RGB channels~\cite{krizhevsky2012imagenet}. For example, slightly altering a word (token) in a sentence may dramatically change its meaning. Therefore, commonly used DA methods for text data are based on synonym replacement from handcrafted ontology word similarity~\cite{kobayashi2018contextual}, inevitably limiting their scope of application because only a small portion of words have precisely or nearly the same meanings. However, MixDA is general enough to generate augmented textual data.

As the first MixDA method for text data, Guo et al.~\cite{guo2019augmenting} adapt Mixup for NLP on the sentence classification task. They propose wordMixup to interpolate word embeddings and senMixup to interpolate sentence embeddings. Experimental evaluations with several network architectures verify their effectiveness. Another similar work is Mixup-Transformer~\cite{sun2020mixup}, which incorporates Mixup with Transformer~\cite{vaswani2017attention} architecture. An issue for pre-trained language models is the miscalibration of in-distribution and OOD data due to over-parameterization. A regularized fine-tuning approach is developed for improved calibration~\cite{kong2020calibrated}. The synthetic on-manifold examples are produced by interpolating in the data manifold to impose a smoothness constraint for better in-distribution calibration. Besides, the predictions for off-manifold instances are forced to be uniformly distributed to mitigate the over-confidence issue for OOD data. Similar to Manifold Mixup~\cite{verma2019manifold}, TMix ~\cite{chen2020mixtext} takes in two text instances and interpolates them in their corresponding hidden space. Emix~\cite{jindal2020augmenting} generates virtual examples by interpolating hidden layer representations with word embeddings and considers the difference in text energies of the samples. Existing embedding-based methods ignore one crucial property: language-compositionality, i.e., a complex expression is built from its sub-structures. A compositional DA approach called TreeMix~\cite{zhang2022treemix} is developed to decompose sentences into their constituent sub-parts by constituency parsing tree. It then mixes them to generate new data, increasing the diversity and injecting compositionality into the model.

\subsection{Sequence-to-Sequence}
Another critical task for text data is the sequence-to-sequence (Seq2Seq) task, which converts sequences from one domain (e.g., sentences in English) to sequences in another domain (e.g., the same sentences translated to French).

SeqMix~\cite{guo2020sequence} is an effective MixDA algorithm for sequence-to-sequence tasks. It generates samples by softly mixing input-output sequence pairs with two binary combination vectors to encourage the neural model's linear behavior. Another approach with the same name is proposed in work~\cite{zhang2020seqmix}, which instead aims to improve the efficiency of active sequence labeling by expanding the queried data and synthesizing new sequences in each iteration. Specifically, SeqMix mixes queried samples in both the feature and target spaces. It guarantees plausibility via a discriminator that computes the perplexity scores for all the constructed candidates and returns the ones with low perplexity. For machine translation, MixDiversity~\cite{li2021mixup} is presented to generate different translations for the input sentence by linearly interpolating it with different sentence pairs sampled from the training corpus during decoding. Another work is multilingual crossover encoder-decoder (mXEncDec)~\cite{cheng2022multilingual}, which interpolates instances from different language pairs into joint "crossover examples" to encourage sharing input and output spaces across languages. To handle the input perturbation issue in machine translation, AdvAug~\cite{cheng2020advaug} is designed to minimize the vicinal risk over virtual sentences sampled from two vicinity distributions, of which the crucial one is a novel vicinity distribution for adversarial sentences that describes a smooth interpolated embedding space centered around observed training sentence pairs. To improve the cross-lingual transfer ability, X-Mixup~\cite{yang2022enhancing} is proposed to mix the representation of the source and target languages during training and inference to accommodate the representation discrepancy in the neural networks. For speech-to-text translation, Speech-TExt Manifold Mixup (STEMM)~\cite{fang2022stemm} is proposed to Mixup the representation sequences of both audio and text modalities, take both unimodal speech sequences and multimodal mixed sequences as input to the translation model in parallel, and regularize their output predictions with a self-learning framework.

\subsection{Graph Neural Networks}
\label{Sub_Graph}
Graph neural networks (GNNs) have gained significant attention recently. While existing MixDA methods have shown promise in various domains, their direct application to graph data may not be suitable. This is primarily because it is challenging to determine how synthetic nodes should be connected to the original nodes through constructed edges while preserving the underlying graph topology.

To handle this problem, GraphMix~\cite{verma2021graphmix} is presented to train an additional fully connected network (FCN) to augment data, and only node features are fed into the FCN. Compared to conventional GNNs that only aggregate information from the neighbors in a few hops due to the over-smoothing issue, GraphMix conducts interpolation on randomly selected nodes, breaking the neighborhood limitation of GNNs and thus improving performance. Motivated by GraphMix, Graph Mixed Random Network Based on PageRank (PMRGNN)~\cite{ma2022graph} expands neighborhood size for the random walk based graph neural networks. To combine both feature and structure information, NodeAug~\cite{xue2021node} is proposed with three variants: the first is NodeAug-I, which only mixes nodes' features and ignores the topology; the second is NodeAug-N, which derives neighbor aggregation of virtual nodes by randomly choosing neighbors of two combined nodes with probability $\lambda$ and $(1-\lambda)$, and then appending edges from these selected nodes; the last is NodeAug-S which scales all related source features accordingly, i.e., adding edges from all neighbors of the combined node to the mixed node. Exploiting the similarity between the neighborhood and receptive field sub-graph, \citet{wang2021mixup} present Mixup-based methods for node and graph classification tasks. Two-branch graph convolution is exploited in the node classification task to interpolate the irregular graph topology and mix their receptive fields. The interpolation is conducted on the aggregated representations from the two branches after each convolution layer. As for graph classification, Mixup performs in the semantic space. Another work, GraphMixup~\cite{wu2021graphmixup}, is designed for class-imbalanced node classification with $3$ sub-modules: (i) feature Mixup performs in a constructed semantic relation space; (ii) two context-based self-supervised strategies are developed to model global and local structure information and conduct edge Mixup; and (iii) a reinforcement Mixup scheme adaptively determines the number of combined nodes. $\mathcal{G}$-Mixup~\cite{han2022g} generates synthetic graphs by interpolating sampled graphons in the Euclidean space, a generator estimated for each class. To create more plausible samples, Graph Transplant~\cite{park2022graph} explicitly uses the node saliency information to guide the selection of sub-graphs and the label generation. Besides, mix-based schemes are also utilized to directly mix node embeddings for better performance in few-label scenarios~\cite{zhao2022synthetic}.

MixDA serves as an effective means to circumvent the over-smoothing issue by mixing distant nodes. Overall, the most critical point for MixDA on graph learning is how to use structure information to compose new data. Solutions are available to mix in the embedding space and combine with the help of sub-graphs or receptive fields. Moreover, directly creating new nodes and their edges using MixDA in a principled manner remains an open challenge.

\subsection{Other Applications}
\noindent \textbf{Metric Learning.} Metric learning (ML) aims to quantify the similarity between examples and obtain an optimal task-specific distance metric, ensuring that embeddings of the same or similar classes are close while dissimilar ones are pushed far apart. On the one hand, some works (e.g., HDML\cite{zheng2019hardness}) have demonstrated that producing new data points and training with metric learning loss can enhance generalization. However, these methods require additional generative networks for data augmentation, increasing the model size and training cost. On the other hand, the loss function of metric learning is based on two or more instances, sharing the same principle with MixDA approaches. Therefore, integrating mixing training into metric learning is preferred. Embedding Expansion\cite{ko2020embedding} synthesizes new data by mingling two or multiple feature representations. It also exploits negative pairs to learn the most discriminative feature. Similarly, Metric Mix (Metrix)\cite{venkataramanan2022takes} adapts Mixup\cite{zhang2018mixup} and Manifold Mixup~\cite{verma2019manifold} into metric learning. It validates the effect of mixed samples with a novel metric called utilization and explores the regions of the embedding space beyond the training classes to refine the representation.

\noindent \textbf{Point Cloud.} Directly applying MixDA to point cloud data may be challenging due to the lack of one-to-one correspondence between the points of two objects. \citet{chen2020pointmixup} formulate data augmentation on point cloud as a shortest path linear interpolation problem and present PointMixup to construct new samples by assigning an optimal path function for two point clouds, resulting in a linear and invariant interpolation. Another work is PA-AUG~\cite{choi2021part}, which splits the objects into multiple sub-parts according to intra-object partition locations and then takes advantage of several MixDA methods in a partition-based way. To maintain topology information of the point cloud samples, Rigid Subset Mix (RSMix)~\cite{lee2021regularization} constructs synthetic examples by replacing an area of one sample with a shape-preserved part from the other sample. This extraction without distortion uses a carefully designed neighboring function considering the point cloud's unordered structure and non-grid properties, thus preserving the structural information of the point cloud data.

\noindent \textbf{Potpourri.} MixDA methods have also been widely applied in various other fields, such as federated learning~\cite{yoon2021fedmix}, vision-language pre-training~\cite{wang2022vlmixer}, semantic segmentation~\cite{kim2023bidirectional}, opinion mining~\cite{miao2020snippext}, speaker verification~\cite{zhang2022contrastive}, speech recognition~\cite{meng2021mixspeech}, video classification~\cite{yun2020videomix}, meta-learning~\cite{yao2021improving, chen2021metamix}, knowledge distillation~\cite{xu2023computation}, collaborative filtering~\cite{moon2023comix} and factorization machines~\cite{wu2022boosting}.

\subsection{Discussion}
\label{Sub_Discussion}
The exploration of MixDA techniques across various data modalities, including image, text, audio, point cloud, and graph, necessitates an understanding of their suitability relative to the inherent characteristics of each modality.

\noindent \textbf{Image.} The initial developments in MixDA, specifically Mixup and Cutmix, focused on image classification tasks, setting a precedent for their extensive application in image data. Consequently, MixDA techniques are predominantly applied and highly effective across various image-related tasks.

\noindent \textbf{Text.} Text data, with its discrete and sequential nature, presents unique challenges and opportunities for MixDA application. Cutmix-based approaches are effective; they allow for the practical manipulation of text at both the data and embedding levels. For instance, cutting and pasting segments of sentences is straightforward. Conversely, Mixup-based methods at the data level are less effective due to the potential generation of non-existent words. However, these issues can be mitigated by applying Mixup at the embedding level.

\noindent \textbf{Audio.} Audio shares certain analytical similarities with text data, allowing for successfully applying Mixup-based and Cutmix-based MixDA methods. Traditional Mixup, for instance, when applied to audio, results in a perceivable overlay of sounds, akin to overhearing simultaneous conversations, which maintains the comprehensibility of the data.

\noindent \textbf{Point Cloud.} The application of MixDA to point cloud is challenging due to the lack of direct correspondence between objects. While direct application of original Mixup and Cutmix is problematic, adaptations such as mixing at the embedding level or employing transformations like optimal path functions and recombining object parts prove viable.

\noindent \textbf{Graph.} Graphs, representing non-Euclidean structures with crucial topological information, require nuanced approaches for data mixing. The complexity of determining labels for mixed samples necessitates domain-specific knowledge, making embedding-level mixing a preferred method. When mixing at the node level, a significant challenge lies in determining connectivity for newly generated nodes.

MixDA's flexibility makes it applicable across diverse data modalities. While certain modalities pose specific challenges, the core concept of interpolating inputs and targets remains universal. Furthermore, embedding-level mixing presents a consistently effective strategy. Future research will need to address the complexities associated with non-conventional data types such as graph and point cloud, continuing to refine and expand the applicability of MixDA.

\section{Explainability Analysis of MixDA}
\label{Sec_Theory}
Although numerous MixDA methods have been successfully used to solve a range of applications, the underlying reasons for their effectiveness remain unclear. In this section, we systematically review the explainability foundations of MixDA, focusing on three different aspects that explain why mixed samples aid generalization: vicinal risk minimization (VRM)~\cite{zhang2018mixup,lim2022noisy,pinto2022regmixup,mangla2020varmixup}, mode regularization~\cite{carratino2020mixup,liang2018understanding,zhang2021does,park2022unified}, and uncertainty \& calibration~\cite{zhang2022and,thulasidasan2019mixup,zhang2020mix}. We also provide some interpretations of why MixDA works well.

\subsection{Vicinal Risk Minimization}
Supervised learning aims to find out a mapping function $f$ in the hypothesis space $\mathcal{F}$ that models the relationship between the input random variable $\mathbf{X}$ and output random variable $\mathbf{Y}$ following a joint distribution $P(\mathbf{X}, \mathbf{Y})$. To achieve this goal, a loss function $\mathcal{L}$ is defined as the discrepancy between the model prediction $f(\mathbf{x})$ and the ground truth $\mathbf{y}$ for the sample $(\mathbf{x}, \mathbf{y}) \sim P$. An optimization algorithm is required to minimize the average of the loss function $\mathcal{L}$ over the joint distribution $P$ to obtain the optimal function $f^{*}$:
\begin{equation}
\label{Eq_ExpectedRM}
    f^{*} = \arg\min_{f \in \mathcal{F}} \int \mathcal{L}(f(\mathbf{x}), \mathbf{y}) \mathrm{d} P(\mathbf{x}, \mathbf{y}).
\end{equation}
Unfortunately, in most cases, the joint distribution $P$ is unknown. As a remedy, the prevailing practice is to collect some training data $\{(\mathbf{x}_{i}, \mathbf{y}_{i})\}_{i=1}^{N}$, where $N$ is the number of training examples and $(\mathbf{x}_{i}, \mathbf{y}_{i}) \sim P$, and utilize empirical risk to approximate expected risk, which is then minimized to attain the optimum, i.e., \textit{empirical risk minimization} (ERM):
\begin{equation}
     P(\mathbf{x}, \mathbf{y}) \approx P_\psi(\mathbf{x}, \mathbf{y}) = \frac{1}{N} \sum_{i=1}^{N} \rho(\mathbf{x}=\mathbf{x}_{i}, \mathbf{y}=\mathbf{y}_{i}),
\end{equation}
where $\rho(\mathbf{x}=\mathbf{x}_{i}, \mathbf{y}=\mathbf{y}_{i})$ is a Dirac mass centered at $(\mathbf{x}_{i}, \mathbf{y}_{i})$. One of the major concerns for ERM is its generalization performance. When the size of the hypothesis space (measured by the number of the model's parameters) is comparable to or larger than the number of training data $N$, the obtained model via ERM is prone to memorizing training samples and, consequently, performs poorly when encountering new data. The reason is that the support of $\rho(\mathbf{x})$ is a one-point set $\{\mathbf{x}_{i}\}_{i=1}^{N}$, therefore, $P_\psi(\mathbf{x}, \mathbf{y})$ cannot approximate $P(\mathbf{X}, \mathbf{Y})$ exactly. To address this problem, \textit{vicinal risk minimization} (VRM)~\cite{chapelle2000vicinal} is proposed to improve the viability of ERM by replacing the Dirac mass with a vicinity function:
\begin{equation}
\label{Eq_VicinityF}
    P(\mathbf{x}, \mathbf{y}) \approx P_{\mathcal{V}}(\tilde{\mathbf{x}}, \tilde{\mathbf{y}})=\frac{1}{N} \sum_{i=1}^{N} \mathcal{V}(\tilde{\mathbf{x}}, \tilde{\mathbf{y}} \mid \mathbf{x}_{i}, \mathbf{y}_{i}),
\end{equation}
where $\mathcal{V}$ is the vicinity function that gauges the probability of the virtual example $(\tilde{\mathbf{x}}, \tilde{\mathbf{y}})$ appears in the vicinity of the training sample $(\mathbf{x}_{i}, \mathbf{y}_{i})$. In this vein, Mixup and Cutmix can be reformulated as a group of generic vicinal distribution:
\begin{align}
\label{Eq_MixVRM}
    \mathcal{V}_{\text{Mixup}}=\frac{1}{N} \sum_{j}^{N} \underset{\lambda}{\mathbb{E}}
    &[\rho(\tilde{\mathbf{x}}=\lambda \mathbf{x}_{i}+(1-\lambda)  \mathbf{x}_{j}, \tilde{\mathbf{y}}=\lambda  \mathbf{y}_{i}+(1-\lambda)  \mathbf{y}_{j})], \notag \\
    \mathcal{V}_{\text{Cutmix}}=\frac{1}{N} \sum_{j}^{N} \underset{\lambda}{\mathbb{E}}
    &[\rho(\tilde{\mathbf{x}}=\mathbf{M} \odot \mathbf{x}_{i}+(1-\mathbf{M}) \odot \mathbf{x}_{j}, \tilde{\mathbf{y}}=\lambda  \mathbf{y}_{i}+(1-\lambda)  \mathbf{y}_{j})].
\end{align}

\subsection{Model Regularization}
From the perspective of model regularization, MixDA methods aim to minimize the standard empirical risk on transformed data with a class of specific perturbations~\cite{carratino2020mixup}. Inspired by previous analyses of dropout~\cite{wager2013dropout,wei2020implicit}, a regularized objective is derived to specify the regularization effects of blended examples~\cite{carratino2020mixup}. It has been demonstrated that Mixup and Cutmix enjoy benefits similar to dropout~\cite{srivastava2014dropout} and label smoothing~\cite{pereyra2017regularizing}. Meanwhile, this work proposes to interpret them to smooth the Jacobian of the model and upgrade the calibration.  It is found that the decision surface trained with mingled samples is smoother than that obtained by conventional ERM~\cite{liang2018understanding}. Another parallel and independent work~\cite{zhang2021does} demonstrates that training with mixed samples is the same as approximating the regularized loss minimization. Specifically, mix-based approaches directly restrain the Rademacher Complexity~\cite{bartlett2002rademacher} of the underlying model and give concrete generalization error bounds. Therefore, it can address the issue of overfitting to some extent. Besides, Mixed sample data augmentation is proven as a pixel-level regularization exposed on the input gradients and Hessians~\cite{park2022unified}. For example, Cutmix actually tries to regularize the input gradients based on pixel distances. 

\subsection{Uncertainty \& Calibration}
Let us consider the most commonly used calibration metric ECE (expected calibration error):
\begin{equation}
    \text{ECE} = \mathbb{E}_{p \sim P_{\hat{p}}}[|\mathbb{P}(\hat{y}=y \mid \hat{p}=p)-p|],
\end{equation}
where $P_{\hat{p}}$ is the probability distribution of $\hat{p}$ -- the largest item in the model output softmax vector. Following~\cite{zhang2022and}, we assume the model is a Gaussian classifier:
\begin{equation}
    f(\mathbf{x}) = \operatorname{sgn}(\mathbf{\omega}^{\mathrm{T}}\mathbf{x}),
\end{equation}
where $\mathbf{\omega} = \sum_{i=1}^N \mathbf{x}_i y_i / N$. Note that $y_i$ is the scalar label ($y_i \in \{-1, +1\}$), which is different from $\mathbf{y}_{i}$ is a one-hot vector. Given $\mathbf{x}$ and $\mathbf{\omega}$, the prediction is obtained by: 
\begin{equation}
    y = f(\mathbf{x})= \arg\max_{c \in\{-1,1\}} p_c(\mathbf{x}),
\end{equation}
where $p_c(\mathbf{x})$ is the probability of class $c$ and is defined as:
\begin{equation}
    p_c(\mathbf{x})=\frac{1}{e^{-2 c \cdot \mathbf{\omega}^{\mathrm{T}} \mathbf{x} / \mathbf{\sigma}^2}+1},
\end{equation}
where $\mathbf{\sigma}$ is the standard deviation vector. After applying MixDA (taking Mixup as an example), augmented data is $\{\tilde{\mathbf{x}}_{i, j}(\lambda), \tilde{y}_{i, j}(\lambda)\}_{i, j=1}^N$, leading to another classifier:
\begin{equation}
    \mathbf{\omega}_{\text{mix}} = \mathbb{E}_{\lambda \sim P_\lambda} \sum_{i, j=1}^N \tilde{\mathbf{x}}_{i, j}(\lambda) \tilde{y}_{i, j}(\lambda) / N^2, \ f_{\text{mix}}(\mathbf{x}) = \operatorname{sgn}(\mathbf{\omega}_{\text{mix}}^{\mathrm{T}}\mathbf{x}),
\end{equation}
where $P_\lambda$ is the distribution of the mix ratio. The result shows that MixDA helps calibration more when the feature dimension is higher: $\text{ECE}(f_{\text{mix}}) < \text{ECE}(f)$. Besides, the above analysis also holds for semi-supervised learning~\cite{zhang2022and}. 

Experiments on several image classification benchmarks and models demonstrate that mix-trained deep neural networks can significantly improve calibration~\cite{thulasidasan2019mixup}. In other words, the generated softmax scores are much closer to the actual likelihood than the conventional models. Besides, solely mingling inputs or features cannot achieve the same degree of calibration. This suggests that the mix of targets, as a form of label smoothing~\cite{szegedy2016rethinking}, plays a vital role in improving calibration~\cite{thulasidasan2019mixup}. In summary, mixed training effectively mitigates over-confidence in data with noise or from out-of-distribution. Moreover, by incorporating Mixup inference, models can be trained using the original one-hot labels, thereby mitigating the negative impact of the confidence penalty~\cite{wang2023pitfall}.

\subsection{Properties \& Interpretations of MixDA}
MixDA possesses numerous appealing properties that have been extensively surveyed in this work. However, only a few works have explicitly summarized these properties and interpreted their functions in improving the model performance. In this subsection, we aim to bridge this gap.

\begin{itemize}
\item MixDA methods such as Mixup and Cutmix oblige linear behavior between training examples. This property can significantly decrease the probability of oscillations when predicting examples that are \textit{not} from the training distribution. It also facilitates model generalization by averting memorization.

\item As a kind of "local linearity" regularization, MixDA training encourages the decision boundaries to transit linearly between classes and smooths the estimate of uncertainty~\cite{verma2019manifold, venkataramanan2022alignmixup, lim2022noisy}. Through the lens of label smoothing~\cite{szegedy2016rethinking}, manipulating mixing targets with the ratio $\lambda : (1 - \lambda)$ prevents the excessive pursuit of the rigid $0$-$1$ estimation. It, in turn, enables the model to account for uncertainty. These merits also boost the calibration and performance in unbalanced scenarios (e.g., in positive and unlabeled learning~\cite{wei2020mixpul,li2022your}).

\item Training with mixed instances is more stable regarding gradient norms and model predictions, which is crucial for generative models such as GANs~\cite{goodfellow2014generative} and DDPM~\cite{ho2020denoising}. Furthermore, MixDA bears some similarities to adversarial training, as both aim to explore areas outside the data manifold. Additionally, as a typical data augmentation method, MixDA alleviates the data-hungry issue in adversarial training~\cite{schmidt2018adversarially}. 

\item Both Mixup and Cutmix methods are simple enough to be incorporated into existing learning models. More importantly, mix operations are usually data-independent and model-agnostic. Consequently, MixDA methods are generic and applicable to various domains (cf. Section~\ref{Sec_Application}).

\end{itemize}

\section{Discussion}
\label{Sec_Discussion}
In this section, we present findings w.r.t. the current research on MixDA and provide insights into the remaining open challenges in this domain. By doing so, future researchers can pinpoint promising future research directions.

\subsection{Revisiting MixDA}
\label{Sub_FindingsChallenges}
By revisiting current MixDA methods, we have the following important findings.
\begin{itemize}
    \item \textbf{Finding 1: Research attention on Mixup and Cutmix.} Despite the similarities between Mixup and Cutmix, their adaptations focus on different aspects. For Mixup, considerable works aim to adaptively determine the mix ratio $\lambda$ (cf. Section~\ref{Sub_Ratio}) or choose appropriate samples for mixing (cf. Section~\ref{Sub_SampleSelection}). In contrast, existing methods for Cutmix focus on studying how to select the cut patch and its location for pasting (cf. Section~\ref{Sub_Saliency2} and Section~\ref{Sub_Divergence}). This difference stems from the fact that Cutmix was proposed from a local perspective with the constructed rectangle region, while Mixup was designed from a holistic view with a global mix ratio.
    \item \textbf{Finding 2: Tradeoff between plausibility and diversity.} MixDA has undergone an evolutionary process with respect to plausibility and diversity. Initially, vanilla Mixup and Cutmix blindly combined training samples. Later, saliency information was leveraged to guide the combination of mixed examples (cf. Section~\ref{Sub_Saliency1} and Section~\ref{Sub_Saliency2}), aiming to generate plausible data. However, the informative patches are limited, resulting in less diverse generated data. Conversely, overemphasizing diversity can lead to many irrational augmented data. Therefore, future work should focus on achieving a favorable tradeoff between these two factors.
    \item \textbf{Finding 3: Mixing more samples and the mix of MixDA.} Most reviewed methods consider how to mix two training samples, while some researchers have started investigating how to combine more examples in each mix process to increase augmentation diversity (cf. Section~\ref{Sub_Diversity1}). Additionally, combining MixDA methods, also known as the mix of MixDA, has gained popularity and has been demonstrated to be effective in further enhancing model performance (cf. Section~\ref{Sub_Others}).
\end{itemize}

\subsection{Open Challenge Problems}
\label{Sub_Challenges}
Despite the desirable performance of MixDA methods in various applications, several challenges persist in training MixDA models. We outline these challenges as follows:
\begin{itemize}
    \item \textbf{Challenge 1: Distortion in Mix.} MixDA methods inevitably introduce some distortion to the original inputs, potentially causing a mismatch between the vicinal and actual data distribution. While this property benefits robustness, mixed examples also introduce noise into the data manifold. Consequently, a critical challenge in the MixDA domain is how to reduce or leverage this perturbation effectively.
    
    \item \textbf{Challenge 2: Combining multiple DA.} Integrating various data augmentation methods into a pipeline is a natural approach for deep learning methods. However, determining the order of multiple data augmentation approaches poses a challenge. For example, it is widely believed that applying Mixup after rotation yields better results than applying rotation after Mixup. Unfortunately, a principled method and theoretical understanding of optimal ordering remain elusive in the literature.
    
    \item \textbf{Challenge 3: The function of $\lambda$.} The function of the mix ratio $\lambda$ is not well understood. Conventional Mixup and Cutmix strategies determine their value empirically using a Beta distribution. Although some research has developed methods to determine this important coefficient adaptively, few studies elucidate the implications behind this ratio and how to determine its value efficiently.
\end{itemize}

\subsection{Research Opportunity}
\label{Sub_FutureWork}
Finally, we outline several exciting research opportunities for future researchers in MixDA. 

\begin{itemize} 
    \item \textbf{Investigating the relationship between MixDA and regularization.} Further investigation is required to explore the relationship between MixDA and regularization. For example, Mixup training requires a significantly lower weight decay on CIFAR-10~\cite{krizhevsky2009learning}, demonstrating that MixDA has cross-cutting and complementary effects with regularization. Besides, Reformulating MixDA as a form of regularization can benefit both lines of work. For instance, exploring the relationship between MixDA and Lipschitz continuity is a potential research proposal.
    
    \item \textbf{Exploring mix-based test-time augmentation for uncertainty estimation and calibration.} Although mix training has demonstrated effectiveness in uncertainty estimation and calibration, using mix-based test-time augmentation for these purposes has been less explored. For example, generating test-time augmented versions for each testing sample by mixing it with examples randomly sampled from the training data is a straightforward solution. However, maintaining a buffer for all training data may be prohibitive. Therefore, finding prototypes or proxies for training examples is an exciting problem that requires further study.
    
    \item \textbf{Integrating MixDA with the large model.} Large models have recently received significant attention and discussion; therefore, using MixDA for large language models is an exciting research direction. First, the few-shot learning capability of large language models provides new application scenarios for MixDA. By interpolating and mixing a small number of samples, the generalization and robustness of the large language model can be further improved. For example, in the in-context learning paradigm of large language models, exploring whether mixing prompts of different tasks can improve the model's task transfer ability is an exciting research direction. Second, regarding large visual language models, image and text features are mixed to alleviate the differences in multimodal data. Moreover, the rich knowledge learned from the large model can also guide the MixDA process at the semantic level, such as preventing the generation of unreasonable mixed samples. Last, MixDA may help improve the security and fairness of large models. By generating hard examples, Mixup exposes models to more diverse adversarial cases, thereby reducing bias and vulnerabilities. Proper feature mixing is also expected to mitigate model bias in tasks involving sensitive attributes.
    
    \item \textbf{Applying MixDA to time series and reinforcement learning tasks.} Extending MixDA methods to time series or reinforcement learning tasks, such as generating scenarios for autonomous driving, presents a pressing challenge due to temporal dependence and the cumulative error issue. A possible avenue is interpolating perceptions from adjacent timestamps to provide more fine-grained information for downstream tasks.

    \item \textbf{Exploring mixing more than two examples.} The field of combining more than two examples in MixDA is still under-explored. This area of research can potentially significantly increase the diversity of augmented data. Iteratively executing Mixup and Cutmix to combine multiple samples can improve the diversity of augmented data and leverage the advantages of Mixup's global view and Cutmix's local perspective.
    
    \item \textbf{Identifying the limits of MixDA.} Identifying the limits of MixDA methods is essential for understanding existing approaches and designing improved ones. For example, when applying MixDA to graph learning, the generated soft labels for the newly constructed nodes can exacerbate the under-confidence issue in GNNs.
\end{itemize}

\section{Conclusion}
\label{Sec_Conclusion}
Data augmentation has consistently been an important research topic in machine learning and deep learning. In this survey, we systematically review the mix-based data augmentation methods by providing an in-depth analysis of techniques, benchmarks, applications, and theoretical foundations. First, we introduce a new classification for MixDA methods. In this context, we present a more fine-grained taxonomy that categorizes existing MixDA approaches into different groups based on their motivations. Then, we thoroughly review various MixDA methods while recapping their advantages and disadvantages. Besides, we comprehensively survey more than eight applications of MixDA. Furthermore, we provide theoretical examinations of MixDA through the lens of VRM, model regularization, and uncertainty \& calibration while explaining the success of MixDA by examining its critical properties. Finally, we summarize our important findings regarding the trends in MixDA research, present the main challenges in existing studies, and outline potential research opportunities for future work in this field.
\bibliographystyle{ACM-Reference-Format}
\bibliography{0_Main.bbl}

\end{document}